# A self-learning framework for adaptive free-form pipe routing in aeroengines

WANG CaiCheng[1], WANG ZiLi[1,2*], ZHANG ShuYou[1,2], XIANG YongZhe[1], LI ZheYi[1] & TAN JianRong[1,2]

[1] State Key Laboratory of Fluid Power and Mechatronic Systems, Zhejiang University, Hangzhou, 310027, China;

[2] Engineering Research Center for Design Engineering and Digital Twin of Zhejiang Province, Zhejiang University, Hangzhou, 310027, China

*Corresponding author: ziliwang@zju.edu.cn

Accepted for publication in SCIENCE CHINA Technological Sciences

**Abstract**

Pipe routing in conventional aeroengine design remains a complex and time-consuming manual task. Although considerable research has been devoted to automating the routing of constant-curvature pipes, the growing demand for free-form pipes introduces new challenges. The dynamic nature of design environments and diverse layout rules further increase the optimization complexity. To tackle these challenges, this study proposes a self-learning pipe routing (SLPR) framework for the automated design of free-form pipes. The framework represents pipe paths using cubic non-uniform B-spline curves to flexibly explore complex spatial domains. A discrete potential energy lookup map is proposed to encode the diverse layout rules. This map enables continuous path exploration with instantaneous rule evaluation through an efficient indexing function. For path optimization, SLPR employs the learning-based proximal policy optimization algorithm. Through iterative interaction with the environment, the SLPR agent refines its routing policy and accumulates reusable knowledge. This knowledge enables the agent to adapt to design modifications through fine-tuning, avoiding the computational overhead of retraining from scratch. Comparative tests reveal that SLPR effectively mitigates redundant segments existing in constant-curvature routing. In static environments, SLPR outperforms diverse algorithmic baselines in terms of path length, rule compliance, path complexity, and computational efficiency. In dynamic environments, the fine-tuned SLPR achieves faster convergence to superior solutions compared to retrained models. These results confirm that SLPR provides an efficient, intelligent, and robust solution for free-form pipe routing, with strong potential for real-world engineering deployment.

## 1 Introduction

Recent advancements in the aerospace industry have driven a higher demand for efficient pipe routing to achieve lightweight, precise, and sustainable aeroengine design. As the foundation for all other detailed designs, pipe routing holds significant research value and practical applications. Typically, pipe routing in an aeroengine requires the precise arrangement of hundreds of pipes while avoiding interference [1]. This complexity renders it a nondeterministic polynomial-time hard (NP-hard) problem. Currently, pipe routing is primarily accomplished manually with the assistance of computer-aided design (CAD) software, a process that can be time-consuming, inefficient, and labor-intensive [2]. The working hours on pipe routing generally occupy more than 50% of the total detail-design person-hours [3]. These high costs have spurred a surge of research into automated pipe routing (APR) [4].

The APR problem can be classified as a path planning problem [5], focusing on determining the shortest, collision-free paths for pipes connecting endpoints within narrow, three-dimensional spaces. Traditional APR approaches often utilize constant-curvature pipes, with pipe paths represented as simple polylines. However, as aeroengine designs become increasingly complex, the available space within the engine is significantly constrained, making constant-curvature pipe routing inefficient due to excessive material use and space occupancy. This shift has driven the adoption of free-form pipes [6], whose axes are represented by spline curves, allowing more flexible obstacle avoidance and reduced pipe length [7].

The high design flexibility inherent in free-form pipe routing introduces new challenges for APR research, with the primary conflict stemming from the trade-off between high-precision environmental modeling and efficient rule modeling [8, 9]. Environment modeling involves defining the layout space using either a discrete or a continuous representation [10]. While continuous space representation offers the geometric freedom essential for optimizing free-form designs, it introduces a formidable challenge for rule modeling. Collision detection is computationally intensive in continuous space due to the presence of irregularly shaped obstacles [11]. Furthermore, pipe routing must adhere to various constraints related to functionality, installation, aesthetics, reliability, and maintenance. Most of these layout rules are fuzzy or ambiguous, making them difficult to express efficiently in a continuous domain. In contrast, discrete space representation divides the environment into discrete units, providing computational convenience and simplifying the encoding of layout rules. Nevertheless, the flexibility for path exploration is limited by the inherent grid resolution.

Beyond the trade-offs in environment and rule modeling, complex and dynamic design environments also challenge the adaptability of route search methods. Over the years, various algorithms have been developed to find optimal layouts in complex environments with multiple constraints. Traditional search algorithms employ point-to-point traversal or sampling to determine the optimal path between the start and target points [12]. Although efficient for specific 2D searches, these algorithms often struggle in large or complex 3D environments. Heuristic-based optimization techniques, inspired by genetic principles or the collective behaviors of organisms, offer advantages such as high parallelism, self-organization, and ease of implementation, making them the most frequently used methods in route search [13]. However, these methods tend to be computationally intensive, prone to getting stuck in local optima [14], and sensitive to the selection of initial parameters [15]. Additionally, both traditional and heuristic-based algorithms lack learning capabilities, requiring a fresh search from scratch for each new task or environment. In contrast, reinforcement learning (RL) algorithms offer a more adaptive, learning-based approach [16]. RL methods allow agents to learn through interactions with the environment and generalize learned behaviors to unseen scenarios [17, 18]. Consequently, RL approaches are well-suited for pipe routing in aeroengines, where frequent design modifications often require extensive adjustments to existing routes. However, such methods are rarely applied to pipe routing.

In summary, existing pipe routing frameworks still face significant limitations in environment modeling, rule modeling, and layout search for free-form pipe design. Most methods either adopt a discrete spatial representation for ease of rule definition or a continuous spatial representation to enhance path-finding flexibility. Moreover, many existing approaches rely on non-learning algorithms, which lack the ability to adapt to dynamic design environments. These challenges are also interdependent and collectively hinder the development of efficient free-form pipe routing. In particular, environment and rule modeling directly influence the performance of the subsequent layout search algorithms, as the routing process requires iterative interactions with the environment for collision detection and fitness evaluation.

To address the aforementioned limitations, this study proposes a self-learning framework for free-form pipe routing in aeroengines. This framework links path exploration in a continuous space with rule evaluation in a discrete space. Specifically, cubic non-uniform B-spline (NURBS) curves are employed to represent continuous pipe geometries, providing the flexibility necessary for lightweight and compact designs. To facilitate the efficient evaluation of candidate paths, a discrete potential energy lookup map is developed offline. An efficient indexing function is proposed to connect the continuous and discrete spaces, allowing for instantaneous fitness assessment during the routing process. The path planning is then implemented using the proximal policy optimization (PPO) algorithm [19], a learning-based method that supports adaptive decision-making under dynamic environmental conditions. The key contributions of this study are as follows:

1) Existing constant-curvature pipe routing methods result in excessive space occupancy and material waste. In this regard, an APR method for free-form pipe routing using NURBS curves is developed to meet the lightweight, precision, and sustainability demands of modern aeroengine design.

2) A potential energy lookup map is constructed in an offline process to encode the diverse pipe layout rules. An efficient indexing function is designed to map continuous coordinates to discrete potential values, seamlessly integrating continuous-space exploration with discrete-space evaluation. This hybrid approach enables real-time rule assessment of the path while preserving the complete geometric freedom inherent in a continuous design space.

3) The PPO algorithm is utilized to optimize the pipe routing paths through continuous interaction with the environment, allowing the agent to learn the flexible obstacle avoidance policy. By integrating transfer learning, the agent further adapts to dynamic changes in the layout environment, enhancing the robustness of the routing framework.

# 2 Related work

## 2.1 Modeling of environment and rules

Environment modeling methods in pipe routing are generally classified into discrete and continuous representations. In discrete modeling, the layout space is partitioned into finite elements. For instance, the raster method divides the layout space into uniform cubic grids, where obstacle-containing grids are assigned a value of one and free grids a value of zero [20]. A pipe path is then defined by a start point, an end point, and a sequence of grids connecting them. This method is particularly convenient for orthogonal pipe routing and is widely used in shipbuilding [21] and factory layouts [22], where orthogonal pipes are preferred. In aeroengine pipe routing, where the space between the engine casing and the nacelle forms a rotational annular space, fan grids are typically used to discretize the layout space [23]. However, the inherently grid-based nature of discrete representations limits the geometric freedom essential for free-form pipes. Although B-spline curves can subsequently fit discrete paths, this post-processing step introduces deviations from the original path, necessitating additional collision checks and evaluations [24].

In contrast, continuous space modeling provides the infinite degrees of freedom necessary for optimal free-form pipe design, enabling agents to explore any point in space. However, this flexibility introduces significant complexity in modeling layout rules. Collision avoidance remains a fundamental principle in pipe routing. Although mesh-based collision detection provides the highest geometric accuracy, it is computationally prohibitive due to the large number of nodes in irregularly shaped obstacles [25]. To alleviate this computational burden, approximation methods such as bounding boxes and voxel-based representations are commonly employed to simplify collision detection [26]. Bounding boxes, such as axis-aligned bounding boxes, spheres, and oriented bounding boxes, provide approximate collision detection by enclosing objects within simple geometric shapes [27]. For more intricate geometries, octree-based adaptive and non-uniform voxelization is employed to recursively partition the space into eight subregions, thereby enhancing collision detection efficiency [28]. Beyond obstacle avoidance, pipe routing must also account for layout rules that encode spatial preferences regarding the relative positioning of pipes and surrounding components. For instance, design rules may require pipes to be placed as close as possible to obstacles and routing boundaries [29]. In discrete space, such preferences are commonly represented using potential fields, where favorable regions are assigned low potential values and unfavorable ones are assigned high values [30]. However, in continuous space, the agent evaluates a unique, unseen location at each decision step, necessitating real-time computation of potential energy values. This requirement imposes a substantial computational load, thereby limiting optimization efficiency.

To address these limitations, we propose a hybrid modeling strategy that combines the flexibility of continuous space with the computational efficiency of discrete evaluation. The proposed approach generates free-form pipe paths in continuous space while assessing their fitness using a discretized potential field that encodes all relevant layout rules. To efficiently bridge the continuous and discrete domains, we construct a potential energy lookup map, which enables rapid queries of precomputed potential values for any point in continuous space via an indexing function. This hybrid strategy seamlessly integrates continuous-space exploration with discrete-space evaluation, providing an efficient and scalable solution for free-form pipe routing.

### 2.2 Route search methods

Recent years have witnessed significant advancements in route search methods [31]. Broadly, route search methods can be classified into traditional search algorithms, heuristic-based optimization algorithms, and learning-based algorithms. Traditional search algorithms employ point-to-point traversal, such as the Dijkstra algorithm [32] and A* algorithm [33], or sampling methods like RRT [34], to identify optimal routes. Heuristic-based optimization techniques include evolutionary algorithms, such as genetic algorithms [35], and swarm intelligence algorithms, such as particle swarm optimization [36] and ant colony optimization [37]. Although widely applied, these methods are inefficient in large, complex 3D spaces and must restart the search from scratch for each new routing task and new environment. The learning-based algorithms, particularly RL methods, address these limitations by learning and adapting through interactions with the environment. For continuous tasks, policy-based algorithms such as soft actor-critic (SAC) [38] and PPO [19] are commonly used, allowing direct mapping from an environmental state to a continuous action space. RL has been successfully applied in diverse path-planning tasks, such as for autonomous underwater vehicles and UAVs in simple, complex, or dynamically changing environments [39-41]. However, the application of RL in pipe routing remains rare. While PPO, combined with curriculum learning, has been applied to ship pipe routing, it has been confined to orthogonal pipes and discrete space [42]. The potential of PPO still warrants further exploration, particularly for the free pipe routing of aeroengines in continuous space.

## 3 Problem formulation

### 3.1 Aeroengine pipe routing

Aeroengine pipe routing identifies a collision-free path between the casing and the nacelle, connecting the designated start and target points. Pipes in aeroengines are typically classified into two types: constant-curvature bent pipes and free-form pipes. The most commonly used are constant-curvature pipes, which are commonly formed by the rotary draw bending process. As modern aeroengines move towards lightweight, precision, and sustainability, the demand for free-form pipes has increased dramatically. As illustrated in Figure 1, both types of pipe paths can be represented as a sequence of axial points as follows:

$$P=\{P_s,P_0,P_1,\ldots,P_n,P_t\}, \tag{1}$$

where $P_s$ denotes the start point, $P_t$ is the target point, and $P_i=(x_i,y_i,z_i)_{0\le i\le n}$ denotes the connection points to be determined. The axial points represent either bend points for a constant-curvature pipe or control points for a free-form pipe, depending on the pipe type. The curve generated from these axial points forms a collision-free path that adheres to routing constraints. The pipe geometry is then constructed by sweeping a circular profile of diameter *d* along this curve.

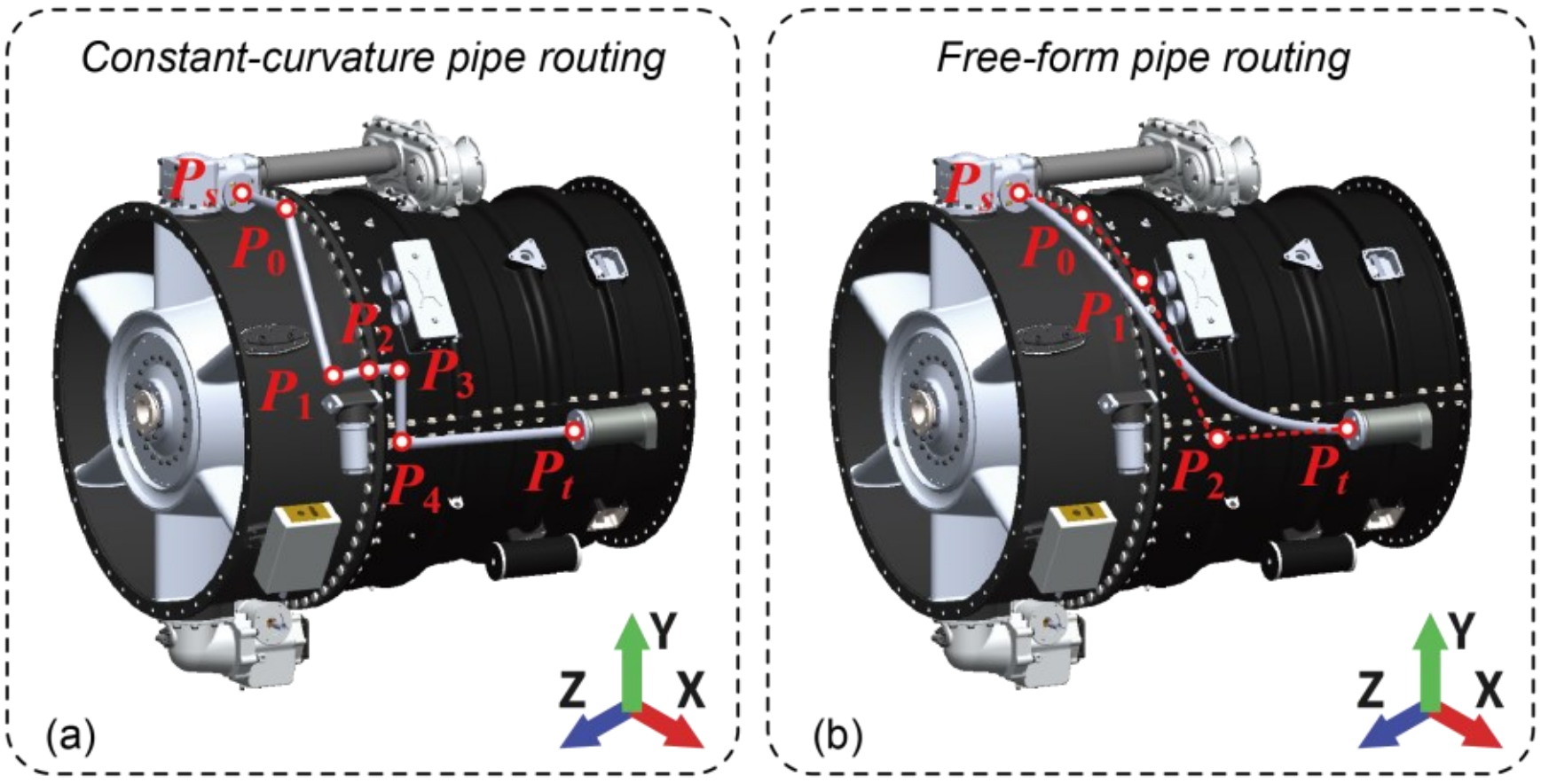

**Figure 1** (Color online) Constant-curvature pipe routing and free-form pipe routing in aeroengine. (a) Constant-curvature pipe routing; (b) Free-form pipe.

Given a series of axial points, a parametric curve can generate a continuous path for the free-form pipe. As one of the most widely used industry standards, NURBS curves are employed in path descriptions due to their precise analytical representation and local tuning capabilities [43]. The NURBS curve can be mathematically expressed as:

$$C(u) = \frac{\sum_{i=0}^{n} N_{i,p}(u) w_i P_i}{\sum_{i=0}^{n} N_{i,p}(u) w_i}, u \in [0,1], \tag{2}$$

where $u$ is the curve parameter, $w_i$ is the weight of the $i$-th control point $P_i$, and $N_{i,p}(u)$ is the corresponding basis function, computed based on the knot vector $U$ and the curve degree $p$. Assume that $U$ is a set of $m$+1 non-decreasing numbers $\{u_0, u_1, \ldots, u_m\}$, the $i$-th basis function $N_{i,p}(u)$ is defined recursively as follows:

$$N_{i,0}(u) = \begin{cases} 1 & \text{if } u_i \le u \le u_{i+1} \\ 0 & \text{otherwise} \end{cases}, \tag{3}$$

$$N_{i,p}(u) = \frac{u - u_i}{u_{i+p} - u_i} N_{i,p-1}(u) + \frac{u_{i+p+1} - u}{u_{i+p+1} - u_{i+1}} N_{i+1,p-1}(u), \tag{4}$$

### 3.2 Reinforcement learning task

The objective of pipe routing is to determine an optimal set of axial control points, which can be formulated as an RL task. In RL, the learner and decision maker are referred to as the agent, while everything it interacts with, comprising all external factors, is referred to as the environment. The RL task is essentially a sequential decision-making problem, commonly represented as a Markov decision process, defined by $M = (S, A, T_s, R, \gamma)$. Here, state $s \in S$ indicates the position and information of the current axial control point within the environment during the routing process. The action $a \in A$ represents the decision to move from the current control point $P_i$ to the next control point. When the agent takes action $a$ in the current state $s$, it transitions to the next state $s'$ according to the state transition function $T_s(s'|s, a)$ and receives a reward determined by the reward function $R(s, a)$. The discount factor $\gamma \in [0, 1]$. measures the importance of future rewards. The goal of the agent is to find an optimal policy $\pi_\phi(a|s)$, which can select the best action $a$ in each state $s$ to maximize the expected return $J(\phi)$:

$$J(\phi) = \mathop{E}_{\tau \sim \pi_\phi} \left[ R(\tau) \right], \tag{5}$$

where $\tau = (s_0, a_0, s_1, \ldots, a_{T-1}, s_T)$ denotes a $T$-step trajectory. Through continuous interactions with the environment, the agent can gradually refine its strategy for determining the axial control points of the free-form pipe.

## 4. Methodology

### 4.1 Environment modeling

The layout space in an aeroengine is defined by the casing and nacelle surfaces, both of which can be considered as revolution surfaces generated by rotating their respective generatrices, $\rho = f_c(z)$ and $\rho = f_n(z)$ around the $z$-axis. As illustrated in Figure 2, the layout space $\Omega$ can be expressed as:

$$\begin{aligned} \Omega = \{(z, \rho, \theta) \in \mathbb{R}^3 \mid & z_{min} < z < z_{max}, \\ & f_c(z) \le \rho < f_n(z), \\ & 0 \le \theta < 2\pi\}, \end{aligned} \tag{6}$$

where $z$, $\rho$, and $\theta$ represent the axes of the cylindrical coordinate system, while $z_{max}$ and $z_{min}$ denote the upper and lower bounds of the layout space on the $z$-axis, respectively.

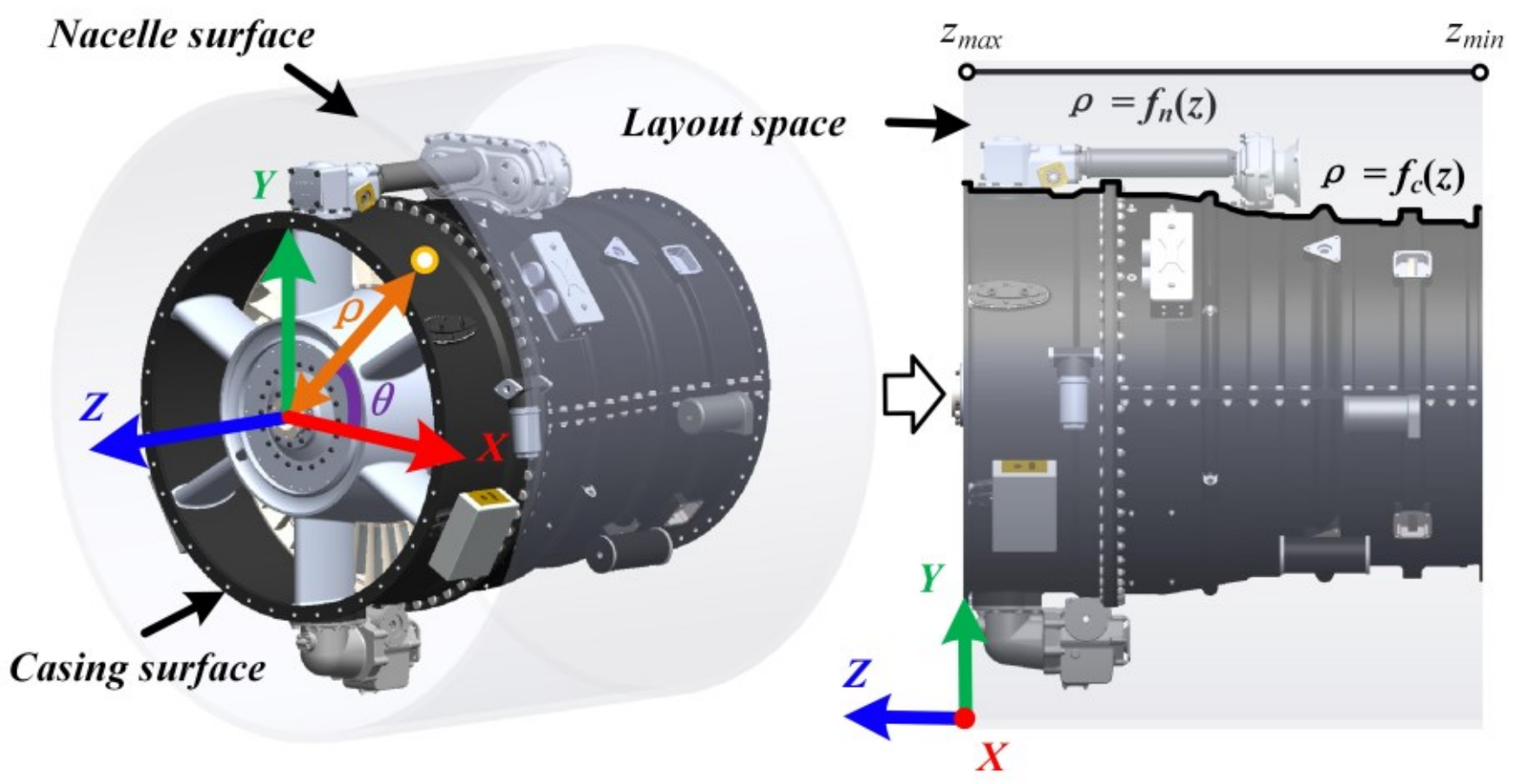

**Figure 2** (Color online) Space of pipe routing in aeroengine.

Within the layout space, various obstacles, such as electrical equipment, fuel tanks, and maintenance areas, restrict the permissible paths for pipes. These obstacles often have irregular shapes, posing challenges for efficient collision detection. Existing research commonly employs bounding boxes to approximate obstacles with simplified geometries. In this study, we employ an octree-based modeling approach to strike a balance between geometric accuracy and computational efficiency [23]. As illustrated in Figure 3 (a), obstacles in CAD software are typically represented as solid geometric models. Point cloud data can be generated by sampling these solid models, which can then be converted into a voxel-based representation using the octree method. The steps of the octree-based modeling method in the cylindrical coordinate system $(z, \rho, \theta)$ are detailed in Figure 3 (b). First, the upper and lower bounds of the obstacle point clouds in the cylindrical coordinate system are used to define the initial modeling space, which serves as the root node of the octree. The initial space is then subdivided into $2^3$ equal subspaces, referred to as the leaf nodes of the octree. Each subspace is classified as an obstacle subspace (marked in black) if it contains at least one point, or as free space (left blank) otherwise. Free subspaces remain unchanged, while each obstacle subspace is treated as a new root node and recursively subdivided until the specified division depth is reached.

The octree data structure employs an adaptive subdivision scheme for obstacle modeling. Unlike uniform grids that partition the entire space evenly, octrees recursively subdivide only those regions that contain geometric entities, while free regions are efficiently represented by larger, undivided nodes. This hierarchical representation significantly reduces memory usage while preserving high geometric fidelity in critical areas. The modeling precision is governed by the division depth, a key parameter that balances accuracy and computational cost. A higher division depth yields a more accurate representation of obstacle geometry but increases computational demands for both storage and traversal. Therefore, it is recommended to select the highest permissible division depth based on the available computational resources and the required modeling accuracy.

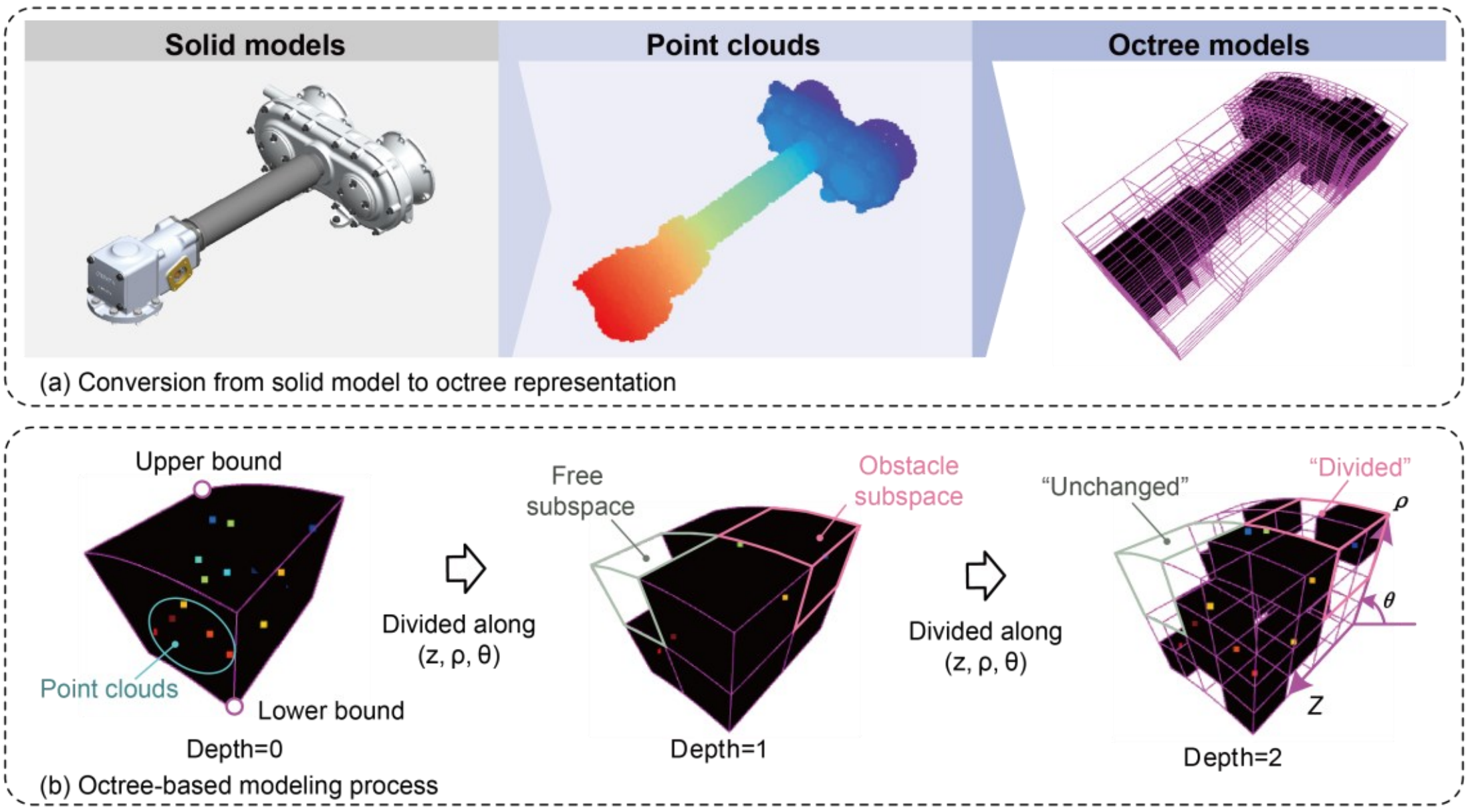

**Figure 3** (Color online) Representation of obstacles through the octree-based modeling approach. (a) Conversion from solid model to octree representation. (b) Octree-based modeling process.

## 4.2 Potential energy lookup map for rule modeling

Pipe routing is governed by numerous layout rules, which can be broadly categorized into layout objectives and layout tendencies, as illustrated in Figure 4. Layout objectives are explicit, quantifiable criteria reflecting the routing goals, such as minimizing pipe length, weight, bend count, and manufacturing costs. Conversely, layout tendencies describe spatial preferences regarding the relative positioning of pipes and surrounding components. These tendencies can be further categorized into deterministic and fuzzy rules, depending on the degree of clarity in their formulation. Deterministic rules provide explicit, quantifiable constraints, such as maintaining a minimum clearance between pipes and obstacles. Fuzzy rules, by contrast, express more flexible and qualitative preferences, such as favoring routing paths that stay close to the casing surface or adjacent accessories.

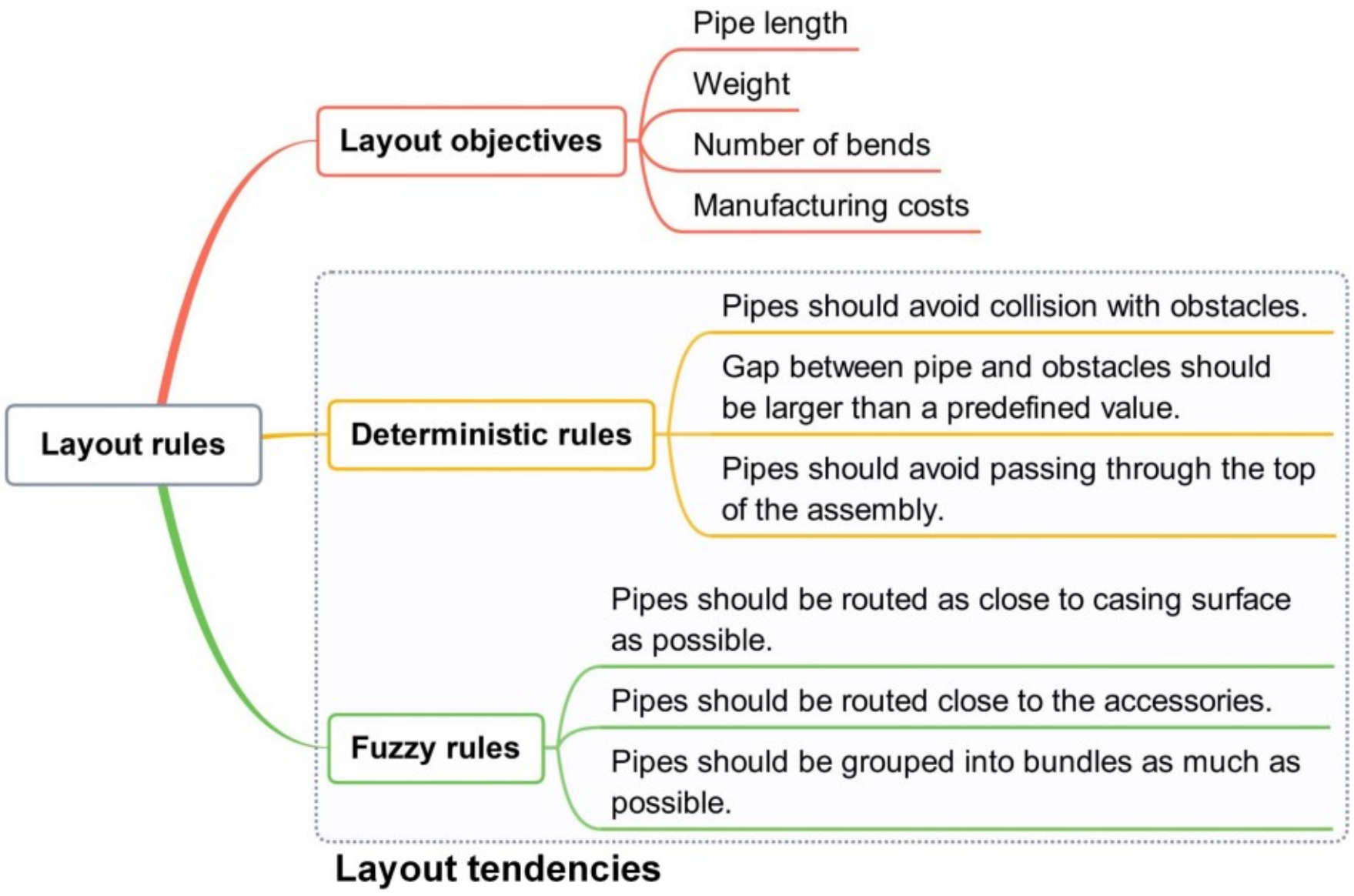

**Figure 4** (Color online) The most common layout rules of pipe routing in aeroengine.

To represent this diverse set of layout rules within a unified mathematical framework, we draw inspiration from the artificial potential field (APF) method, which is widely applied in path and robotics planning [44]. The APF constructs a virtual potential field over the environment to guide a mobile agent safely and efficiently from its starting point to the target. This potential field, denoted by $U$, is typically defined as the superposition of an attractive field and a repulsive field:

$$U = U_{att} + U_{rep}, \tag{7}$$

where $U_{att}$ is the attractive potential that pulls the agent towards the goal, and $U_{rep}$ is the repulsive potential that pushes the agent away from obstacles to ensure collision avoidance. In this study, we adapt and extend the APF method to accommodate the more intricate requirements of pipe routing, including both deterministic and fuzzy rules.

To encode diverse layout tendencies, we utilize potential energy fields to represent spatial preferences. The external objects influencing the pipe layout, such as obstacles, casing surfaces, or previously installed pipes, are collectively referred to as external components (ECs). For each EC, a repulsive potential field is generated in the region where pipe placement is discouraged. The repulsive potential $U_{rep}$ at the point $P$ within this region is defined as:

$$U_{rep}(P) = k_r, \quad P \in Z_r, \tag{8}$$

where $k_r$ is the repulsive gain factor representing the strength of the potential, and $Z_r$ denotes the repulsive influence zone. This zone is the union of two subregions: a collision avoidance zone $Z_{collision}$ and an installation access zone $Z_{access}$. The $Z_{collision}$ ensures a minimum clearance between the pipe and EC, preventing physical interference. It is defined as:

$$Z_{collision} = \{P \in \Omega \mid d(P, \mathrm{EC}) \le r_0 + \delta_{clearance}\}, \tag{9}$$

where $d(P, EC)$ denotes the Euclidean distance between $P$ and the EC, $r_0$ is the pipe radius, and $\delta_{clearance}$ is the required safety clearance. To account for practical considerations such as installation and maintenance, pipes are also discouraged from passing directly above external components. This restriction is enforced via the installation access zone $Z_{access}$, which is defined as:

$$Z_{access} = \{P \in \Omega \mid (z, \theta) \in \mathrm{Proj}_{z\theta}(\mathrm{EC}), \rho > \rho_{\mathrm{EC,max}}\}, \tag{10}$$

where $\mathrm{Proj}_{z\theta}(\mathrm{EC})$ denotes the projection of the EC onto the $z\theta$-plane, and $\rho_{\mathrm{EC,\,max}}$ is the maximum radial coordinate of the EC. The complete repulsive influence zone is then given by:

$$Z_r = Z_{collision} \cup Z_{access}. \tag{11}$$

Any portion of the pipe path entering the impermissible region $Z_r$ incurs a penalty via the assigned negative potential energy, effectively discouraging undesired configurations.

Conversely, an attractive potential field is generated around an EC in regions where pipe placement is preferred. The attractive potential energy $U_{att}$ is defined as:

$$U_{att}(P) = k_a \times \frac{\delta_{outer} - d(P, EC)}{\delta_{outer} - \delta_{inner}}, \quad P \in Z_a, \tag{12}$$

where $k_a$ is the attractive gain factor representing the maximum strength of the field, and the influence zone $Z_a$ is a shell-like region bounded by inner and outer distances $\delta_{inner}$ and $\delta_{outer}$, respectively:

$$Z_a = \{P \in \Omega \mid \delta_{inner} \le d(P, \mathrm{EC}) \le \delta_{outer}\}. \tag{13}$$

The potential energy of $U_{att}$ is maximized at the inner boundary of the $Z_a$ and linearly decreases to zero at its outer boundary.

Since an EC may simultaneously exert both attractive and repulsive influences in different regions, the total potential field $U_j$ generated by the $j$-th EC is expressed as:

$$U_j(P) = \begin{cases} U_{att,j}(P), & P \in Z_{a,j} \\ U_{rep,j}(P), & P \in Z_{r,j} \\ 0, & P \notin Z_{a,j} \cup Z_{r,j} \end{cases}. \tag{14}$$

The overall potential energy at $P$ is then computed as the linear superposition of the potential fields generated by all $m$ ECs:

$$U(P) = \sum_{j=1}^{m} U_j(P). \tag{15}$$

Frequent evaluation of total potential energy is required during the pipe routing process. In the continuous space, pipe routing requires recalculating the potential at each step, a computationally expensive task dominated by repeated distance calculations of $d(P_{a,i}, EC)$. Although leveraging the octree-based modeling simplifies the distance calculation, this remains a significant efficiency bottleneck within the iterative optimization loop.

To overcome the computational burden associated with repeated potential energy evaluations, we propose a strategy based on pre-computation and an efficient lookup mechanism. The entire layout space is discretized into a grid of uniform, fan-shaped cells of size $s$, as illustrated in Figure 5. The potential energy at the center of each cell is precomputed and stored in a three-dimensional array, referred to as the potential energy map $P_e$. Let $C_{i_z,i_\rho,i_\theta}$ denote the center point of the cell indexed by $(i_z, i_\rho, i_\theta)$, the potential value for this cell is computed as:

$$P_e(i_z, i_\rho, i_\theta) = U(C_{i_z,i_\rho,i_\theta}). \tag{16}$$

This one-time operation yields a static potential field representation across the entire design domain, as shown in Figure 5 (c). During the optimization phase, the potential energy of any point $P = (z, \rho, \theta)$ in the continuous space can be retrieved almost instantaneously by mapping its coordinates to the discrete indices ($i_z$,$i_\rho$,$i_\theta$) using the following indexing function:

$$\begin{cases} i_z = \lfloor (z - z_{\min}) / s \rfloor \\ i_\rho = \lfloor (\rho - \rho_{\min}) / s \rfloor, \\ i_\theta = \lfloor \theta \times \rho_{\min} / s \rfloor \end{cases} \tag{17}$$

where $\lfloor\ \rfloor$ denotes the floor function, and ($z_{min}$, $\rho_{min}$) represent the lower bounds of the layout space along the $z$-axis and the $\rho$-axis. The corresponding potential energy is then queried directly via $P_e(P) = P_e(i_z, i_\rho, i_\theta)$. This direct indexing approach eliminates the need for expensive distance calculations within the optimization loop, significantly improving computational efficiency.

The potential energy lookup map is also designed to support sequential multi-pipe layout scenarios by treating previously routed pipes as new ECs within the map. The repulsive and attractive fields generated by the previously routed pipes help prevent inter-pipe collisions and promote optimized pipe bundling, respectively.

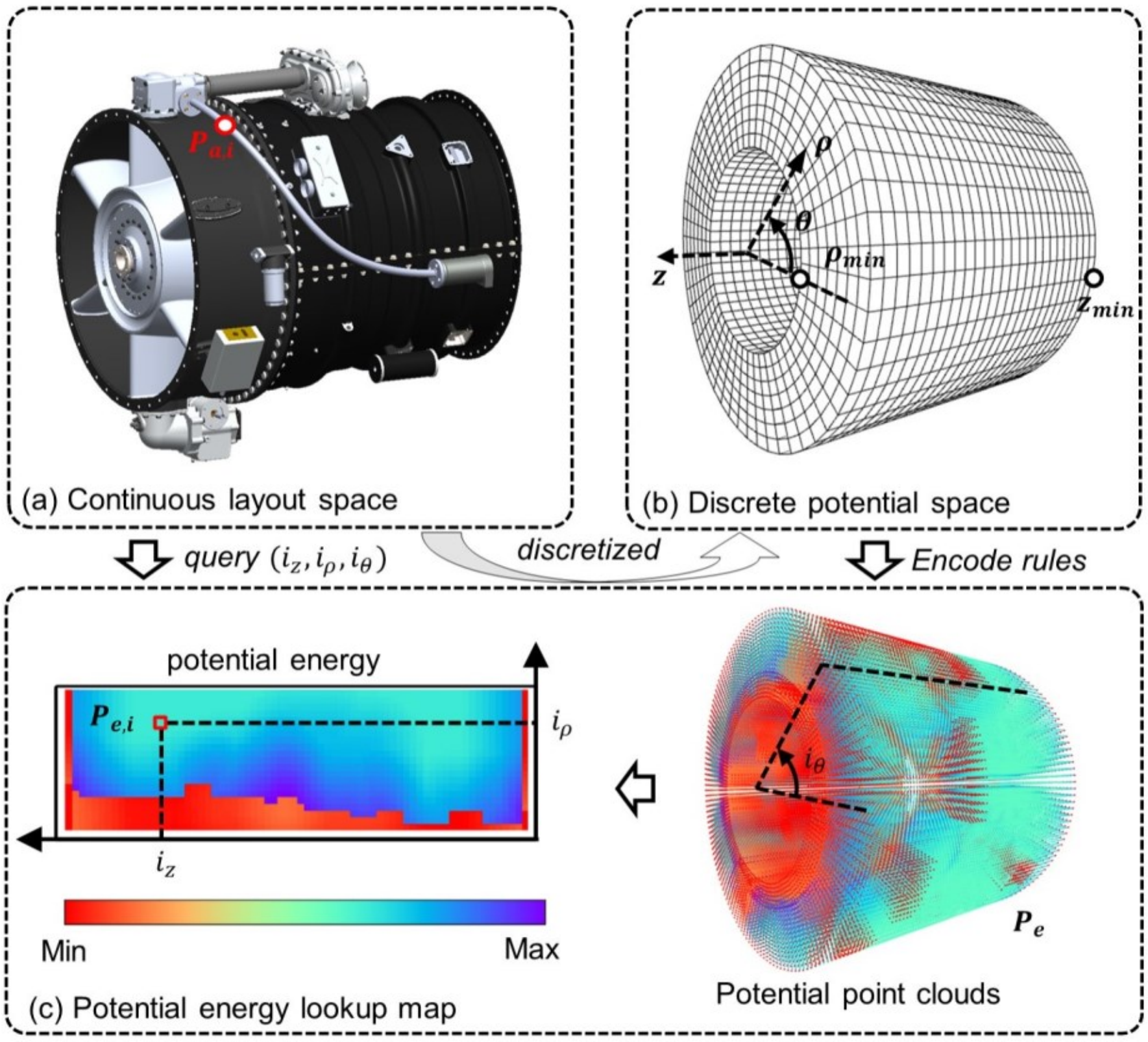

**Figure 5** (Color online) Potential energy lookup map for layout rule modeling. (a) Continuous layout space; (b) Discrete potential space; (c) Potential energy lookup map.

## 4.3 Self-learning pipe routing with NURBS curve

The PPO algorithm is utilized as the core method to optimize the policy $\pi_\phi(a|s)$ of pipe routing. It is also combined with a fine-tuning approach to adapt to dynamic layout environments, as illustrated in Figure 6. PPO is an on-policy, actor-critic algorithm that effectively balances training stability and computational efficiency. In PPO, the actor represents the learned policy that selects actions, while the critic evaluates the policy by estimating the state-value function $V(s)$. The algorithm optimizes a clipped surrogate objective, constraining policy updates to ensure stable training. The loss function of the actor-network is defined as follows:

$$L_{\text{actor}}(\phi) = \mathbb{E}_i\left[\min\left(r_i(\phi)\hat{A}_i, \text{clip}(r_i(\phi), 1-\varepsilon, 1+\varepsilon)\hat{A}_i\right)\right], \tag{18}$$

where $r_i(\phi) = \frac{\pi_\phi(a_i|s_i)}{\pi_{\phi old}(a_i|s_i)}$ is the probability ratio, with $\phi_{old}$ being the policy parameters prior to the update. $\varepsilon$ is a hyperparameter that controls the magnitude of policy updates. The advantage estimate $\hat{A}_i$ is computed using generalized advantage estimation (GAE) and is expressed as:

$$\hat{A}_i = \delta_i + (\gamma\lambda)\delta_{i+1} + \ldots + (\gamma\lambda)^{n-i+1}\delta_{n-1}, \tag{19}$$

where $\lambda$ is a hyperparameter that balances the trade-off between bias and variance, and $\delta_i = r_i + \gamma V(s_{i+1}) - V(s_i)$ represents the temporal difference error**.** The critic loss evaluates the accuracy of the value function in approximating the true state value. A commonly used metric is the mean squared error between the predicted value $V_w(s_i)$ and the target value $V_{target}(s_i)$, formulated as:

$$L_{\text{critic}}(\omega) = \mathbb{E}_t\left[(V_\omega(s_i) - V_{\text{target}}(s_i))^2\right], \tag{20}$$

with *w* being the parameters of the critic network.

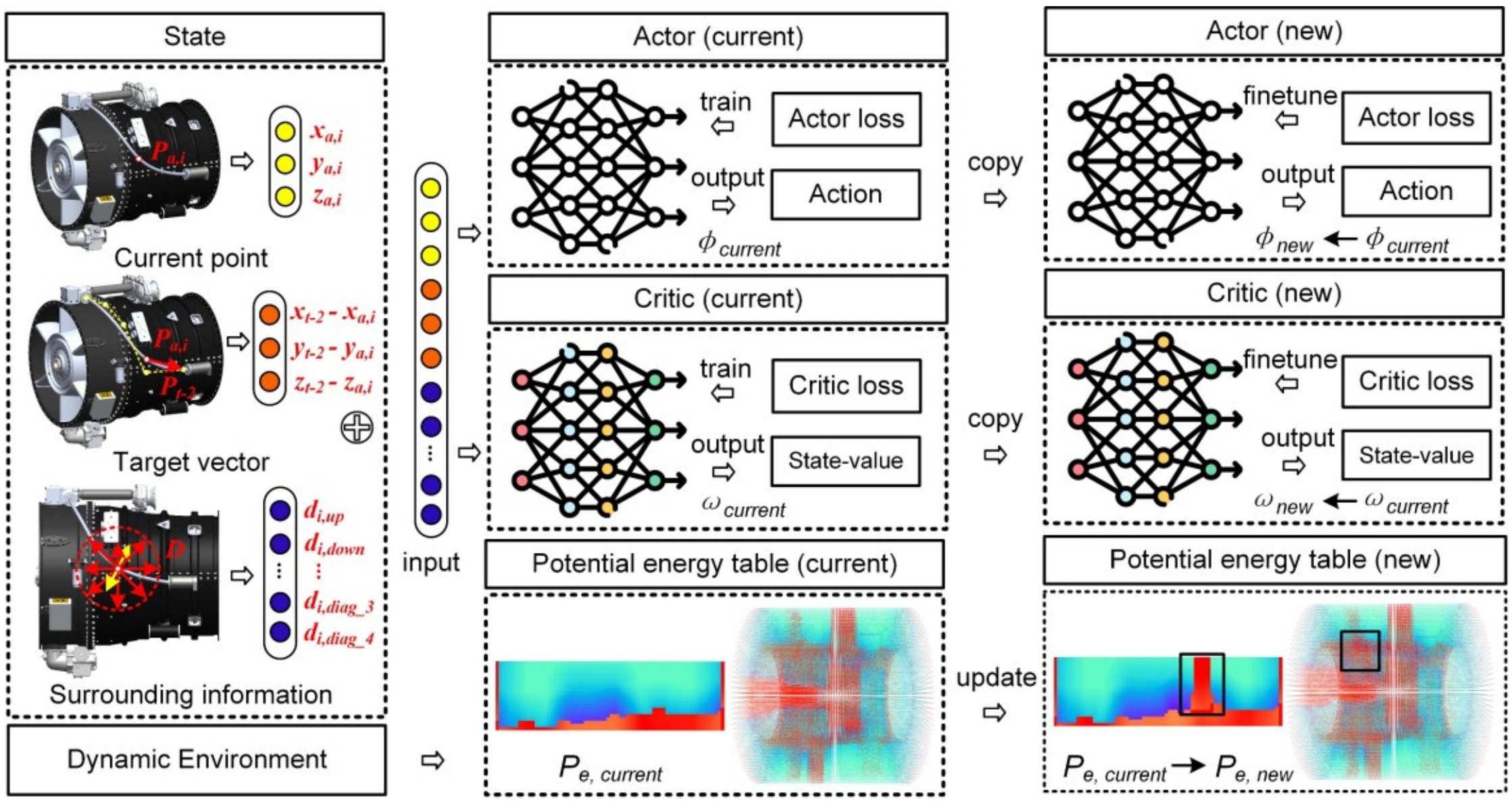

Figure 6 (Color online) Framework of self-learning pipe routing.

The configuration of the state space, action space, and reward function is crucial for the convergence speed and training performance of PPO. As illustrated in Figure 6, the state space of the agent consists of three main parts, represented as:

$$S_i = [P_{a,i}, P_t - P_{a,i}, D_i], \tag{21}$$

where $P_{a,i} = (x_{a,i}, y_{a,i}, z_{a,i})$ denotes the current position of the agent, initialized as $P_s$ at the $i$ = 0. $P_t - P_{a,i}$ represents the layout task information. $D = [d_{i,up}, d_{i,down}, \dots d_{diag_3}, d_{diag_4}]$ captures the agent's surrounding information. This vector comprises the distances to the nearest impassable region along the primary axes (upward, downward, forward, backward, left, right) and four diagonal directions in space. The three components are concatenated and utilized as input to the PPO model.

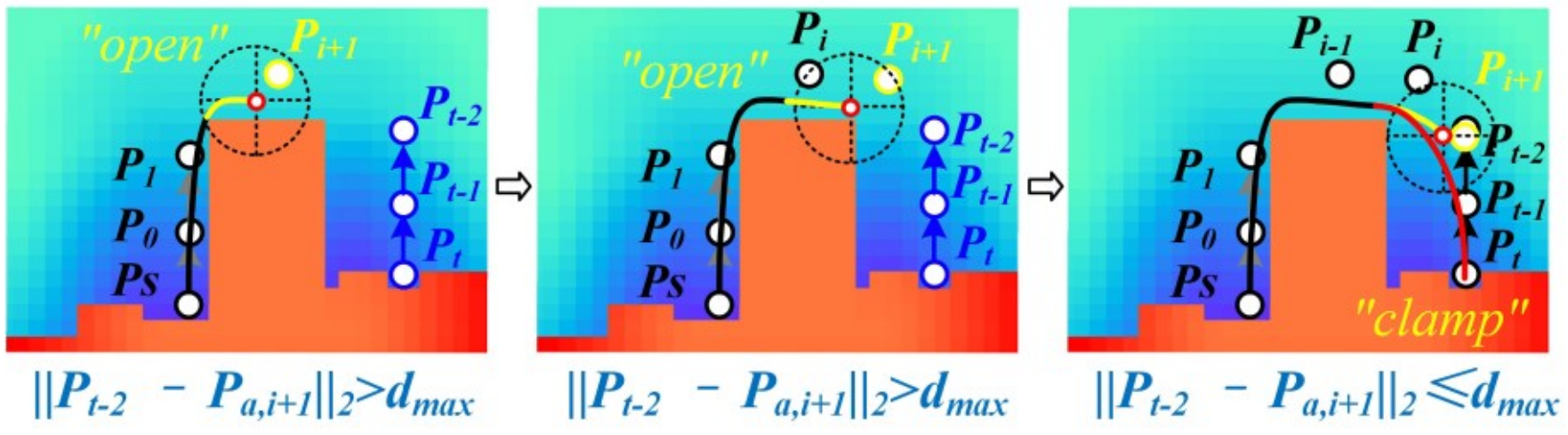

(a) (b) (c)

Figure 7 (Color online) Pipe routing using cubic NURBS curve. (a) Initial layout; (b) Intermediate process; (c) Termination when finding the target.

Cubic NURBS are employed during the PPO search process to generate pipe paths. A minimum of four control points is required for the generation of cubic NURBS curves. $P_0$ and $P_1$ are derived by extending the pipe diameter and twice the pipe diameter along the normal direction of the $P_s$ port, respectively. Similarly, $P_{t-1}$ and $P_{t-2}$ are determined by extending $P_t$ in the same manner, as illustrated in Figure 7. This setting ensures that the pipe path is tangent to the start and target ports. The $P_s$, $P_0$, and $P_1$ are considered as the initial control points with default weights of 1, while subsequent control points are sequentially obtained by defining the action space of the agent as follows:

$$A_i = [\Delta x_{i+1}, \Delta y_{i+1}, \Delta z_{i+1}, w_{i+1}]. \tag{22}$$

Here, $w_{i+1}$ denotes the weight of the next control point $P_{i+1}$. The vector $[\Delta x_{i+1}, \Delta y_{i+1}, \Delta z_{i+1}]$ represents the displacement of $P_{i+1}$ relative to $P_i$ along the $X$, $Y$, and $Z$ axes, respectively. The magnitude of this displacement is constrained by a predefined upper limit:

$$\sqrt{(\Delta x_{i+1})^2 + (\Delta y_{i+1})^2 + (\Delta z_{i+1})^2} \le d_{\max}, \tag{23}$$

where $d_{max}$ defines the maximum allowable distance between consecutive control points. The new control point $P_{i+1}$ is then appended to the list of control points $P$. By manipulating the knot vectors of NURBS, a new segment of an open curve, which is not tangent to the line $P_iP_{i+1}$, is generated at the end of the last curve. The search process continues as long as the distance between the $P_{a,i}$ and $P_{t-2}$ exceeds the predefined threshold $d_{max}$. Once this threshold is met, the target is considered reached. The final points $P_{t-2}$, $P_{t-1}$, and $P_t$ are appended to $P$ with weights of 1 to form a clamp curve tangent to the line $P_{t-1}P_t$, thus seamlessly connecting $P_s$ and $P_t$.

The reward function evaluates the corresponding curve segment whenever a new control point is identified. Five distinct reward functions are designed to prevent abnormal behaviors such as greed, timidity, and recklessness arising from reward sparsity. The first reward function, $R_1$, encourages the pipe path to approach the target point and is expressed as:

$$R_1 = \left\| P_{t-2} - P_{a,i} \right\|_2 - \left\| P_{t-2} - P_{a,i+1} \right\|_2, \tag{24}$$

$R_1$ rewards the agent when it moves closer to the target point compared to its previous position; otherwise, it imposes a penalty. Another reward function, $R_2$, incorporates the pipe length as a critical pipe layout objective and is expressed as:

$$R_2 = -l_{i+1}, \tag{25}$$

where $l_{i+1}$ denotes the length of the new curve segment. To ensure the pipe adheres to layout rules, the potential energy of the curve is also evaluated. Specifically, sampling at equal intervals $\Delta l$ along the curve segments, the potential energy $P_{e,j}$ is calculated at each sampling point. Based on this, two additional rewards, $R_3$ and $R_4$, are defined as follows:

$$R_3 = \begin{cases} \text{mean}(P_{e,j} \mid P_{e,j} \ge 0) - \max(P_e), & P_{e,j} \notin \varnothing \\ 0, & \text{otherwise} \end{cases}, \tag{26}$$

$$R_4 = \text{sum}(P_{e,j} \mid P_{e,j} < 0). \tag{27}$$

Here, $R_3$ defines the reward for points with $P_{e,j} \ge 0$ by averaging their potential energies and subtracting $\max(P_e)$. The mean and maximum operations in $R_3$ ensure its value remains negative, thereby discouraging greedy agent behavior. In contrast, $R_4$ penalizes sampling points that enter impassable regions ($P_{e,j} < 0$). Finally, when the distance between the $P_{a,i}$ and $P_{t-2}$ is less than $d_{max}$, the target point is considered reached, and the agent receives the reward $R_5$, defined as:

$$R_5 = \begin{cases} 0, & \left\| P_{t-2} - P_{a,i+1} \right\|_2 > d_{\max} \\ 1, & \left\| P_{t-2} - P_{a,i+1} \right\|_2 \le d_{\max} \end{cases}. \tag{28}$$

In summary, the total reward function $R$ is defined as a weighted sum of all sub-reward functions:

$$R = \mu_1 R_1 + \mu_2 R_2 + \mu_3 R_3 + \mu_4 R_4 + \mu_5 R_5, \tag{29}$$

where $\mu_1$, $\mu_2$, $\mu_3$, $\mu_4$, and $\mu_5$ are the weights assigned to each sub-reward function. According to the definition of the reward function, the agent's objective is to identify the shortest path connecting the designated start and end ports while satisfying both deterministic layout rules (e.g., obstacle avoidance) and fuzzy layout preferences (e.g., proximity to casing surface), as illustrated in Figure 4.

In the pipe routing process, frequent modifications to engine design often necessitate re-routing the planned pipe, significantly increasing design person-hours. To address this challenge, this study integrates the PPO algorithm with a fine-tuning methodology, as schematically illustrated in Figure 8. When the environment changes, the potential energy map is first updated to reflect the new environment. Subsequently, the parameters of the fully trained actor and critic models, $\phi_{current}$ and $\omega_{current}$, are transferred to the new model. This knowledge transfer preserves the agent's previously learned routing policies, allowing it to bypass a full retraining. Instead, the agent can quickly adapt to the new environment through a brief fine-tuning phase, which employs a smaller learning rate and significantly fewer training episodes.

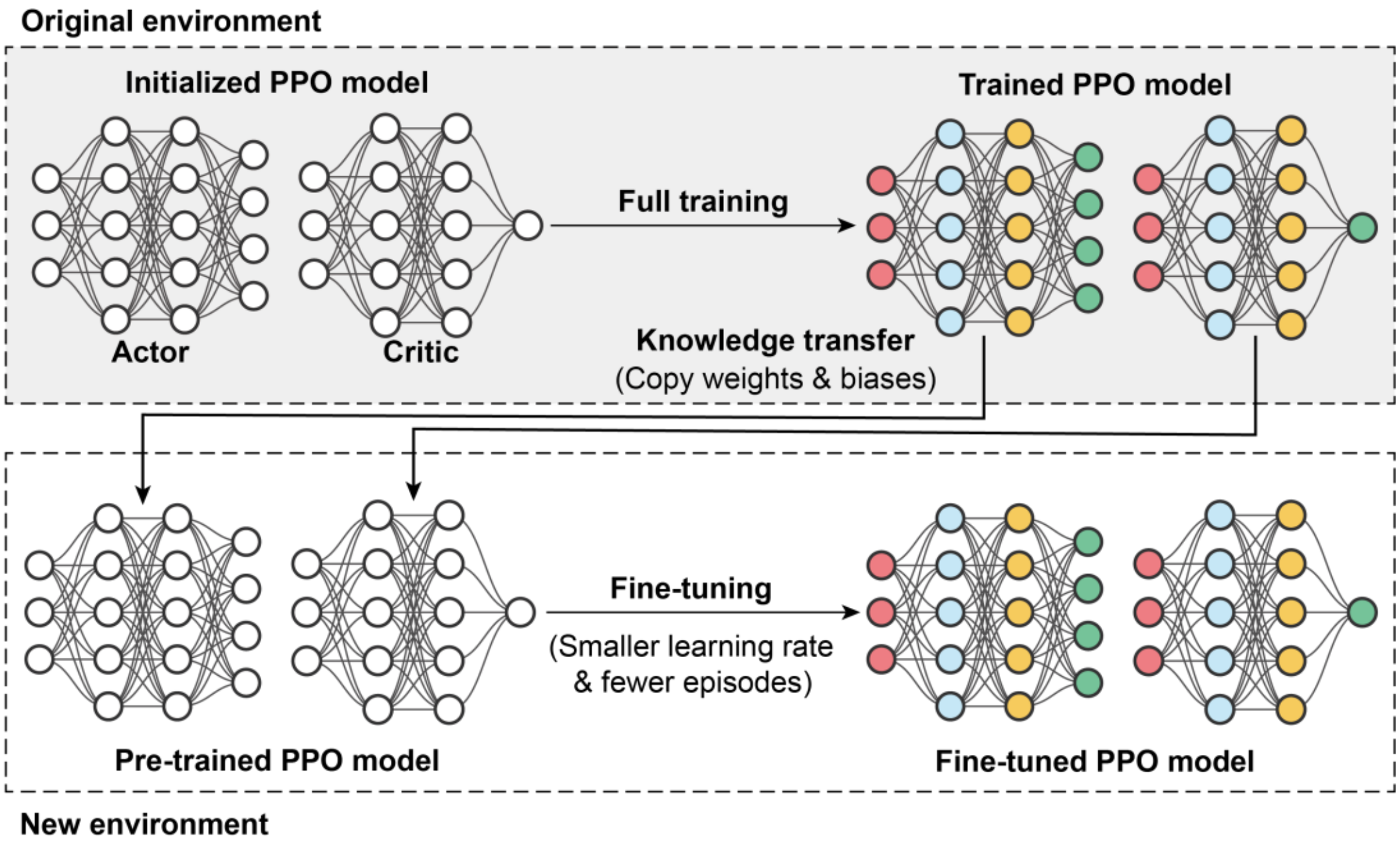

**Figure 8** (Color online) Schematic of SLPR adaptation to new environments via fine-tuning.

# 5 Experiments and results comparison

## 5.1 Experimental setup

To evaluate the performance of the proposed self-learning pipe routing (SLPR) framework, an aeroengine model based on the CAD assembly available on GrabCAD [45] was prepared in SolidWorks. The layout space was defined based on the designed casing generatrix $f_c(z)$ and nacelle generatrix $f_n(z)$, which represent the upper and lower bounds of the layout space, respectively. $f_c(z)$ is a polynomial function with a minimum value of 418.75, while $f_n(z)$ is a constant set to 800. The point clouds of the obstacles were exported to construct the potential energy map $P_e$, with the cell size $s$ of $P_e$ set equal to the pipe diameter.

The terms $k_a$ and $k_r$ represent the strengths of the attractive and repulsive fields, respectively. To encode the hierarchical relationship between hard constraints (avoidance) and soft preferences (guidance) for pipe routing, the framework emphasizes the relative magnitude of these gains rather than their absolute values. Specifically, the repulsive gain is standardized as $k_r = -1$, and the attractive gain is defined relative to it as $k_a = 0.2$. The underlying principle is that the penalty for violating a hard constraint must be substantially greater than the incentive for satisfying a soft preference. In this implementation, the absolute magnitude of the penalty is five times greater than the maximum attractive potential energy. This configuration ensures that entering a repulsive field is strictly prohibited, whereas moving within an attractive field is merely preferred.

The pipe routing experiments were then conducted in a Python-based simulation environment. The SLPR networks were implemented in PyTorch and trained on a cloud platform with a 32 vCPU AMD EPYC 9654 96-Core Processor. The main hyperparameter settings of SLPR are summarized in Table 1, determined through a combination of automated tuning tools [46] and manual refinement.

The reward function weights $[\mu_1, \mu_2, \mu_3, \mu_4, \mu_5]$ in Table 1 are critical for balancing the competing objectives of the routing task. $\mu_1$ is the weight of the process reward $R_1$, which encourages the agent to move toward the target point, mitigating the issue of sparse rewards. An excessively large $\mu_1$ would make the agent overly aggressive in pursuing the target while neglecting other constraints. Since the agent's action range is limited to 100 mm, $\mu_1$ is set to 0.01 to keep $R_1$ within the range [−1, 1], thereby stabilizing the training. $R_5$ represents the final layout objective and thus requires a relatively large weight to exert a significant influence on the total return. Given that the maximum number of steps per episode is 20, $\mu_5$ is set to 10 so that the final discounted $R_5$ has a comparable magnitude to $R_1$. $R_2$ denotes the length penalty, with its weight $\mu_2$ defined relative to $\mu_1$. This penalty encourages the agent to select shorter paths while moving toward the target. Since this objective is less critical than $R_1$, $\mu_2$ is set to 0.002, which is an order of magnitude lower than $\mu_1$. $\mu_4$ is the weight of the violation penalty. To maintain consistency with the repulsive field strength $k_r = -1$, $\mu_4$ is also set to 1. An excessively large $\mu_4$ would make the agent overly cautious. The weight $\mu_3$ is set relative to $\mu_4$ and was determined to be 0.05 through parameter tuning. Since $R_3$ is defined as a negative value to prevent greedy agent behavior, an excessively large $\mu_3$ would also cause the agent to stagnate.

**Table 1** Summary of model hyperparameter configurations in SLPR

| Parameter | Settings |
|---|---|
| Total training episodes | 5000 |
| Maximum steps per episode | 20 |
| Hidden layers of each MLP | 2 |
| Neurons of each hidden layer | 256 |
| Discount factor $\gamma$ | 0.9 |
| GAE factor $\lambda$ | 0.98 |
| Clip ratio $\varepsilon$ | 0.2 |
| Weights [$\mu_1$, $\mu_2$, $\mu_3$, $\mu_4$, $\mu_5$] | [0.01, 0.002, 0.05, 1, 10] |
| Optimizer type | Adam |
| Learning rate of actor | $1\times10^{-4}$ |
| Learning rate of critic | $5\times10^{-4}$ |
| Judgment threshold $d_{max}$ | 100 mm |
| Sampling interval $\Delta l$ | 5 mm |

Three comparative experiments were conducted across different scenarios, including a comparison with constant-curvature pipe routing, a layout performance comparison with other algorithms, and a layout performance test in a dynamic environment. The simulation results are analyzed in detail below.

## 5.2 Comparison with constant-curvature pipe routing

Aeroengine pipe routing has traditionally relied on constant-curvature pipes, which often result in redundant segments, increased path lengths, and suboptimal space utilization. In contrast, the SLPR framework proposed in this study is specifically designed to optimize free-form pipe paths represented by NURBS curves, offering a more effective solution for the complex annular layout spaces characteristic of aeroengines. To evaluate the geometric advantages of free-form pipe routing, we conducted a comparative study against a baseline method based on traditional constant-curvature routing. The baseline paths were generated using the interactive routing tools provided in SolidWorks, a widely used industrial CAD platform.

Three distinct pipe routing tasks were designed, each with a pipe diameter of 27 mm. A comparison of layout lengths is summarized in Table 2. The results demonstrate that the free-form pipes consistently produced shorter routing paths across all test scenarios, with reductions of approximately 16.4%, 19.4%, and 9.4% for the three respective pipe configurations. Visualizations of the resulting layouts are shown in Figure 9. These results illustrate that the free-form paths generated by SLPR not only minimize pipe length but also reduce geometric complexity, particularly in scenarios involving long-distance routing and obstacle avoidance within confined annular spaces. Constant-curvature pipes inherently lack flexibility due to the discrete nature of their bending segments. Although increasing the number of bends can improve routing flexibility, it simultaneously introduces greater manufacturing complexity. In contrast, the use of cubic NURBS curves in SLPR enables the generation of smooth, continuous paths with $C^2$ continuity, a critical requirement for ensuring both structural integrity and manufacturability.

**Table 2** Comparison of layout lengths for different pipe types.

| Length (mm) | Constant-curvature pipe | Free-form pipe (SLPR) |
| --- | --- | --- |
| Pipe-1 | 1404.66 | 1174.02 |
| Pipe-2 | 1812.20 | 1460.82 |
| Pipe-3 | 1267.43 | 1148.50 |

Furthermore, existing industrial software typically provides semi-automated routing tools that rely on predefined search heuristics and require substantial manual intervention. Consequently, the quality of the final layout is highly dependent on the engineer's experience and expertise. The proposed SLPR framework addresses this limitation by offering a data-driven, self-learning alternative. By continuously interacting with the environment, SLPR autonomously learns optimal routing policies, thereby enhancing layout quality while reducing dependence on manual tuning and domain expertise.

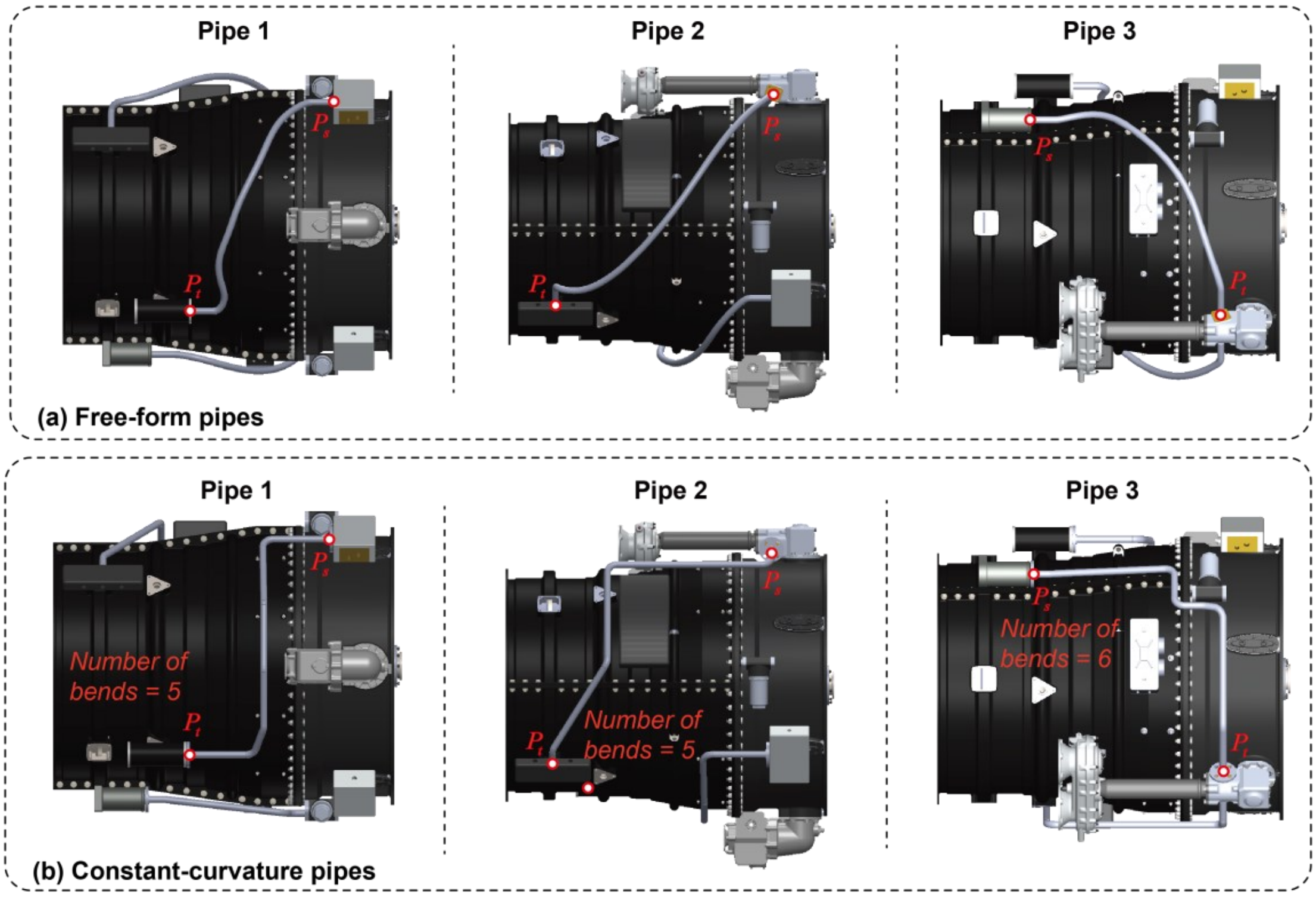

**Figure 9** (Color online) Visualization of the layout schemes for different pipe types. (a) Free-form pipe routing; (b) Constant-curvature pipe routing.

## 5.3 Comparative analysis against diverse algorithmic paradigms

In this subsection, three representative algorithmic paradigms were selected to evaluate the performance of the proposed SLPR through comparative experiments in a static design environment. These methods include SAC, B-spline RRT* (BSRRT*), and quantum particle swarm optimization (QPSO), corresponding to RL algorithms, sampling-based search algorithms, and heuristic optimization algorithms. The details of each method are described below.

### (1) SAC

The SAC is a state-of-the-art RL method for solving complex control tasks in continuous action spaces [38]. It combines value-based and policy-based techniques, employing the maximum entropy framework to maximize both the expected reward and policy entropy. The entropy regularization coefficient is initially set to 0.12 and is automatically adjusted during training to strike a balance between exploration and exploitation. Other parameter settings for SAC are consistent with those outlined in Section 5.1.

### (2) BSRRT*

The RRT* algorithm searches for a path by extending an exploration tree from the root node to the target. It enhances path quality through rewiring and reselecting the optimal parent node. The BSRRT* algorithm, a recently proposed variant of RRT*, is designed to directly identify control points for quadratic B-spline curves in smooth path planning [47], making it a suitable benchmark for comparison with the proposed SLPR. BSRRT* also integrates an expanded candidate and goal-biased strategies to accelerate convergence. However, its application has been limited to two dimensions. In this experiment, the cubic B-spline curves were used instead to accommodate the 3D layout environment.

### (3) QPSO

PSO is a widely used population-based intelligent optimization algorithm inspired by the collective behavior of organisms in nature. QPSO introduces the quantum mechanics theory into the PSO algorithm, reducing the number of tunable parameters while enhancing the optimization performance [36]. Within QPSO, particles are represented as a collection of coordinates and weights for all control points, with the number of control points fixed at 10. The optimal solution is found through cooperation and interaction among particles. The swarm size is set to 30, and the number of iterations is set to 1000, ensuring a balance between efficiency and performance.

To enable a comprehensive performance comparison, ten distinct pipe routing tasks were designed within the aeroengine model, each with a pipe diameter of 19 mm. These tasks cover scenarios such as short-distance unobstructed layouts, long-distance layouts, and obstacle avoidance in narrow spaces. The evaluation metrics included the pipe length, the average potential energy of sampling points, the number of rule-violating sampling points, the number of successful tasks, the number of control points, and the computational time. These metrics evaluate performance in terms of pipe length reduction, adherence to layout rules, path complexity, and computational efficiency.

The comparison of average performance metrics for SLPR and other algorithms across the ten layout tasks is presented in Table 3, with the best results highlighted in bold. To ensure fairness, all methods were executed within the same computational environment, utilizing a shared potential energy map $P_e$ for path evaluation. The construction of the map is a one-time, offline pre-computation that requires 1.79 minutes. Once constructed, this map provides near-instantaneous potential energy lookups for all algorithms during their execution. The results in the table highlight SLPR's superior overall performance. In terms of computational efficiency, SLPR achieved an average online training time of 6.88 minutes, which is substantially faster than the optimization times required by SAC (11.20 minutes), BSRRT* (11.68 minutes), and QPSO (11.82 minutes). This rapid convergence demonstrates the effectiveness of the PPO algorithm within our framework. Moreover, this efficiency is achieved without compromising the quality of the solution. SLPR successfully identifies the shortest collision-free pipe paths that fully comply with the layout rules, achieving a total pipe length of 670.48 mm, zero violation points, and a 100% task success rate. Meanwhile, SLPR exhibits the lowest path complexity, with an average of five control points, which is advantageous for the manufacturability of free-form pipes.

**Table 3** Average performance of SLPR and several comparative methods on free-form pipe routing

| Metrics | SLPR | SAC | BSRRT* | QPSO |
|---|---|---|---|---|
| Length (mm) | 670.48 | 733.73 | 980.15 | 1495.07 |
| Potential energy | -0.16 | -0.40 | -0.21 | -0.15 |
| Num of violation points | 0 | 25.80 | 1.5 | 0 |
| Successful times | 10 | 9 | 10 | 10 |
| Num of control points | 5 | 7 | 9 | 10 |
| Agent training time (min) | 6.88 | 11.20 | 11.68 | 11.82 |
| Construction time for $P_e$ (min) | | 1.79 | | |

A detailed comparison of the results for each tube, including pipe length, number of violation points, average potential energy, and number of control points, is presented in the line chart in Figure 10. The RL algorithms, SLPR and SAC, outperform BSRRT* and QPSO in optimizing pipe length. Nevertheless, it is noted that the obstacle-avoidance performance of the SAC algorithm is unstable, with violation points occurring in half of the layout tasks, leading to the worst performance in terms of average potential energy. The BSRRT* algorithm performed well on tasks where the start and end points were close, such as pipes 2, 4, 7, and 10; however, it resulted in a violation on task 6. For pipes 2 and 10, BSRRT* identified the simplest pipe paths that met the requirements with a few control points. However, for longer layout tasks, the generated paths contained too many control points, resulting in overly complex pipe paths with redundant segments. This is because the RRT* algorithm generates the tree by randomly sampling points, which can lead to locally congested paths or unnecessary detours during tree expansion. The QPSO algorithm yields the best results in terms of average potential energy and successfully finds collision-free pipe paths across all tasks, excelling in adherence to layout rules. However, while QPSO can find feasible paths, they are not always optimal, particularly for pipes 3, 7, 8, and 9, where path redundancy is significant. This issue arises from the unique coding pattern of the heuristic algorithm, in which particles are represented by concatenating the coordinates and weights of all control points. Since the dimensions of the particles are fixed, the number of control points must be determined in advance. Fewer control points may result in paths that fail to bypass obstacles, while too many control points increase the difficulty of particle search and impair the optimization performance of QPSO. Therefore, selecting an appropriate number of control points according to the layout task is essential for QPSO.

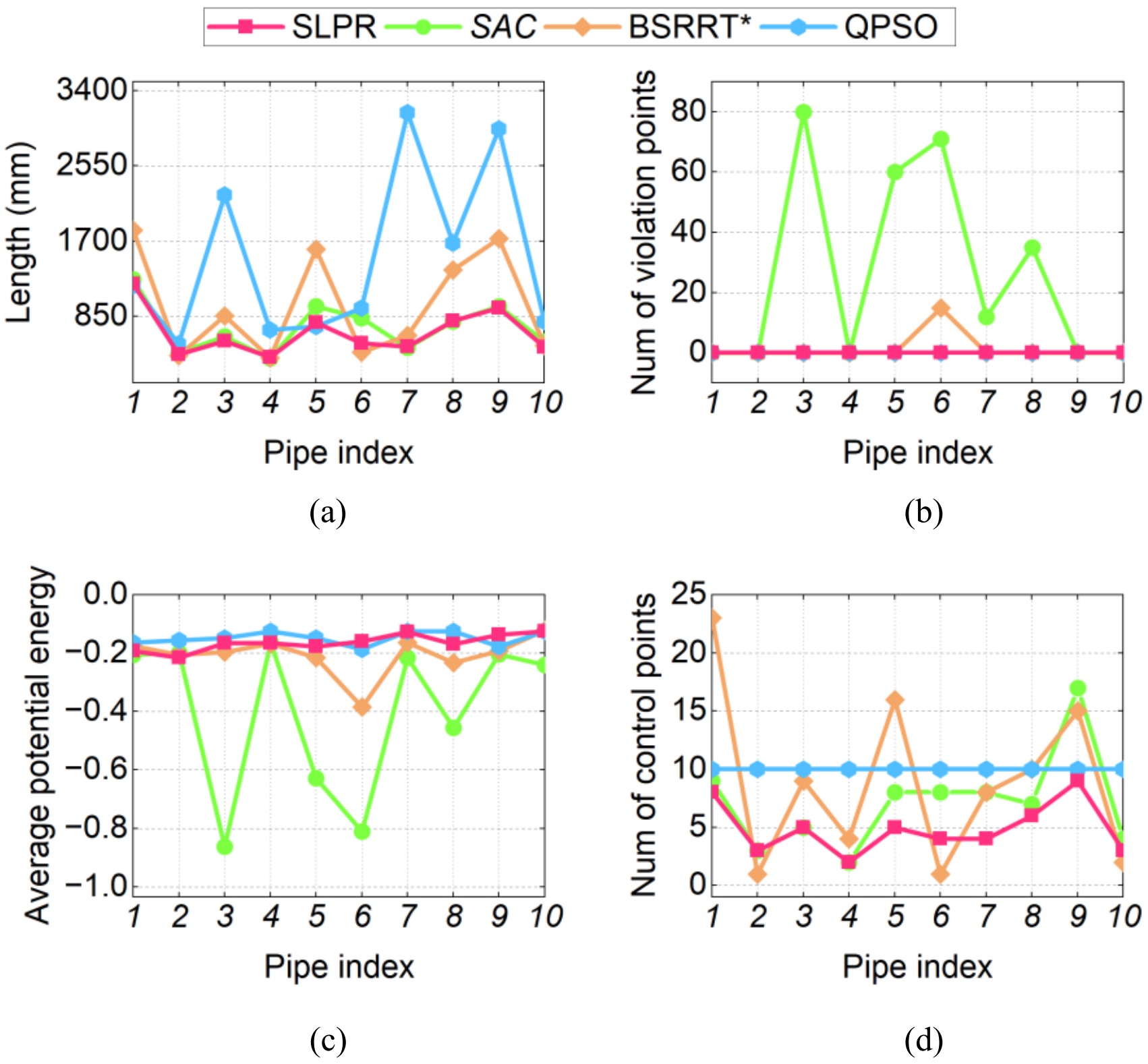

**Figure 10** Performance comparison for each pipe. (a) Length; (b) Num of violation points; (c) Average potential energy; (d) Num of control points.

The visualization of pipe routing results for SLPR is given in Figure 11. As shown in the figure, for short-distance unobstructed layout tasks, including pipes 2 and 4, SLPR generates short and straight paths to minimize pipe length. For long-distance layout tasks, such as pipes 1 and 3, SLPR provides smooth, curved paths that conform to the annular surface of the aeroengine. In scenarios requiring obstacle avoidance, such as pipes 5, 6, and 7, SLPR successfully routes pipes around obstacles while avoiding passage over the top of obstacles, which adheres to the layout rules and demonstrates the effectiveness of the potential energy map in standardizing the pipe routing. Finally, for layout tasks in narrow spaces requiring obstacle avoidance, including pipes 8, 9, and 10, SLPR still provides feasible, collision-free paths, showcasing its robust search capabilities.

The results demonstrate that the proposed SLPR outperforms SAC, BSRRT*, and QPSO in pipe routing capabilities. SLPR identifies the shortest collision-free pipe path that satisfies the layout rules with the fastest speed and exhibits the lowest path complexity, facilitating easier pipe forming. These advantages make SLPR applicable to pipe routing and manufacturing in real-world scenarios.

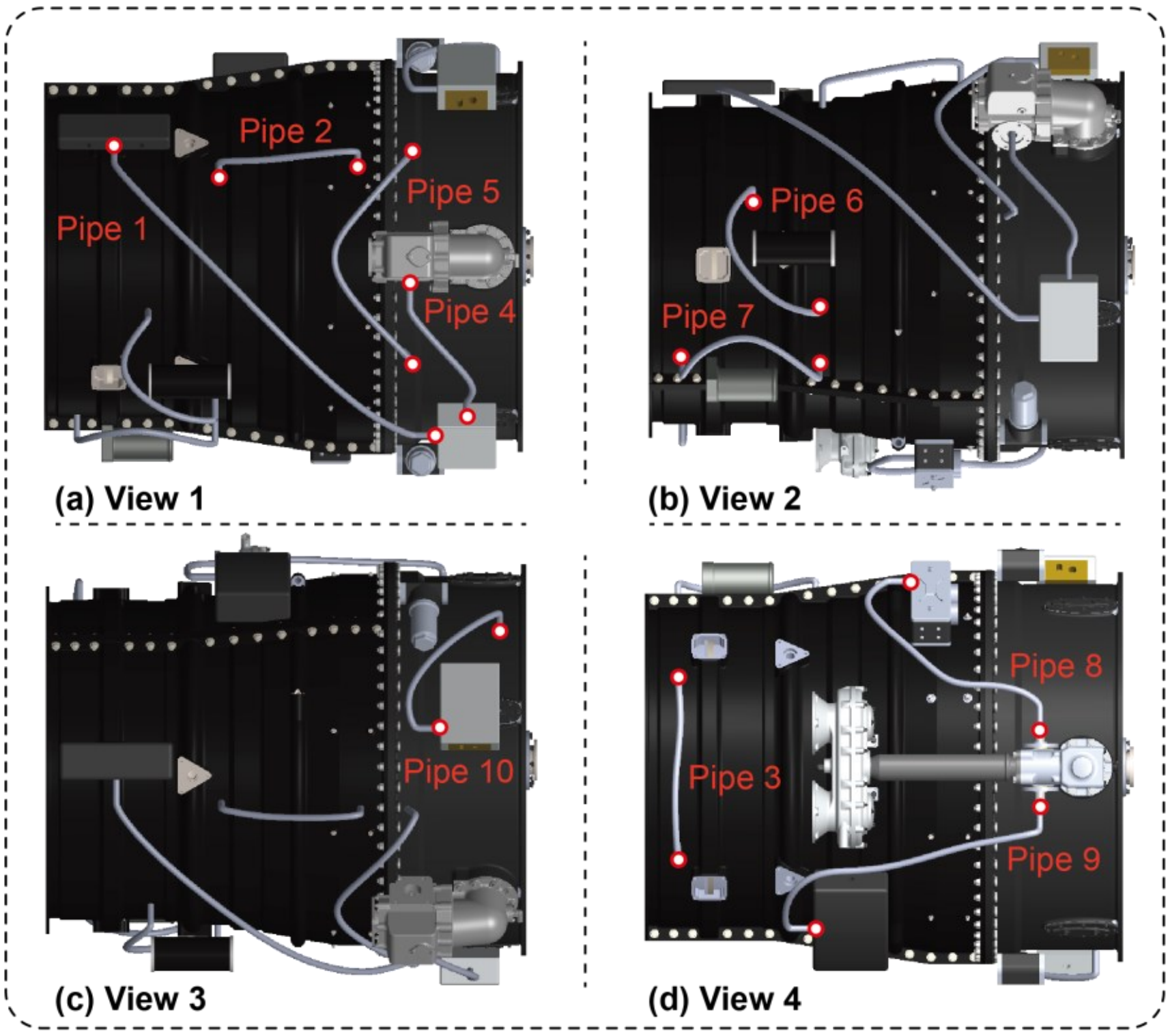

**Figure 11** Results of the free pipe routing for SLPR. (a) View 1; (b) View 2; (c) View 3; (d) View 4.

### 5.4 Adaptation to design modifications via fine-tuning

Real-world aeroengine design is an iterative process that often involves frequent modifications, such as the addition, removal, or relocation of components. These changes may lead to collisions with previously planned pipe layouts or degrade previously optimal solutions to suboptimal ones. In such cases, traditional search and heuristic algorithms must restart the entire route from scratch, significantly increasing design time and computational effort. In contrast, the learning-based SLPR is designed to retain the prior knowledge, allowing it to adapt to new layout environments through fine-tuning. To demonstrate the adaptive mechanism and computational efficiency of SLPR, two representative scenarios were constructed to emulate common design modifications, as shown in Figure 12. The first scenario introduces a new obstacle that causes a collision with the previously planned pipe, simulating the addition of a new component to the layout. The second scenario involves modifying an existing obstacle, representing the relocation or resizing of a component.

To enable successful transfer, the potential energy map is first updated to reflect the environmental changes. The old SLPR model, previously trained in the original environment, is then directly deployed in the modified environment, as shown in Figure 12 (b). The results indicate that the agent can respond adaptively to environmental changes. In the "New obstacle" scenario, the agent successfully avoids collisions, although it fails to reach the target point. In the "Modified Obstacle" scenario, the agent adjusts the path slightly to accommodate the new obstacle boundaries, yielding a shorter pipe length. These outcomes demonstrate that the agent can leverage its prior knowledge to adapt to environmental changes, providing a strong initial solution that can be rapidly optimized through fine-tuning.

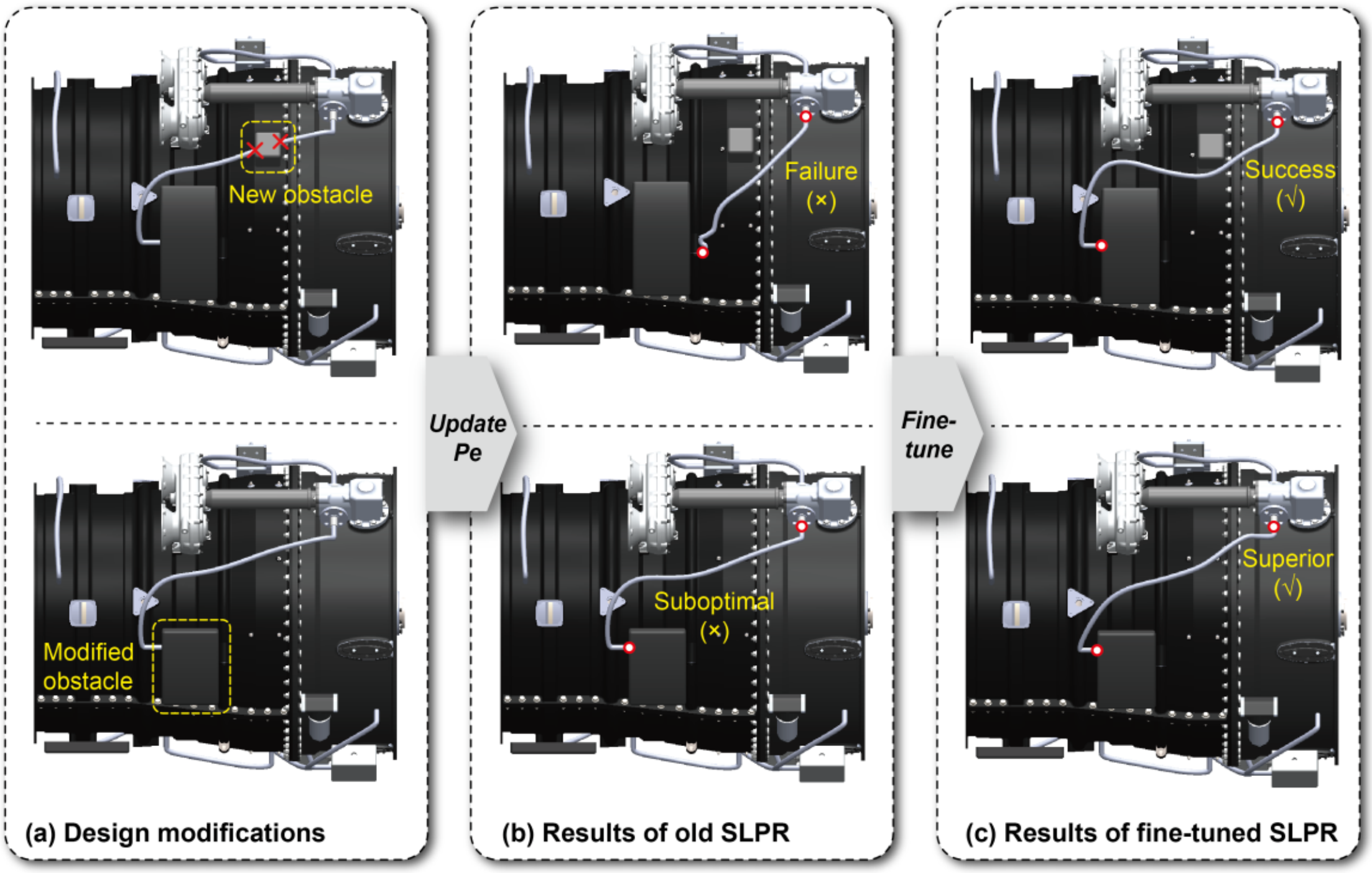

**Figure 12** Pipe routing in a dynamic environment. (a) Design modifications; (b) Results of old SLPR; (c) Results of fine-tuned SLPR.

The previously trained SLPR model was further fine-tuned in the modified environment using the updated potential energy map while maintaining all parameters consistent with those described in Section 5.1, except for the total number of training episodes, which was set to 100. The visualization results of the fine-tuning are presented in Figure 12 (c). After fine-tuning, the pipe successfully bypasses the new obstacle and reaches the target point. In contrast, the pipe in the “Modified Obstacle” scenario achieves a shorter length. Furthermore, the iterative curves of the maximum return achieved by the SLPR during the fine-tuning and retraining are presented in Figure 13. The results indicate that the fine-tuned model maintains a high return from the beginning of training. In the "Modified Obstacle" scenario, the fine-tuned model converged to its maximum return (7.61) in just 32 episodes. In contrast, the model retrained from scratch required 2,257 episodes to reach a similar performance level. This represents a convergence speed-up of approximately 70 times. In the "New Obstacle" scenario, the advantage is even more pronounced. The fine-tuned model achieved a superior solution (max return of 6.57) in only 49 episodes. The retrained model, even after 3,694 episodes, only reached a suboptimal return of 6.03, failing to match the quality of the fine-tuned solution. These results suggest that leveraging prior knowledge not only accelerates convergence to new environments but can also lead to superior final solutions compared to retraining from scratch.

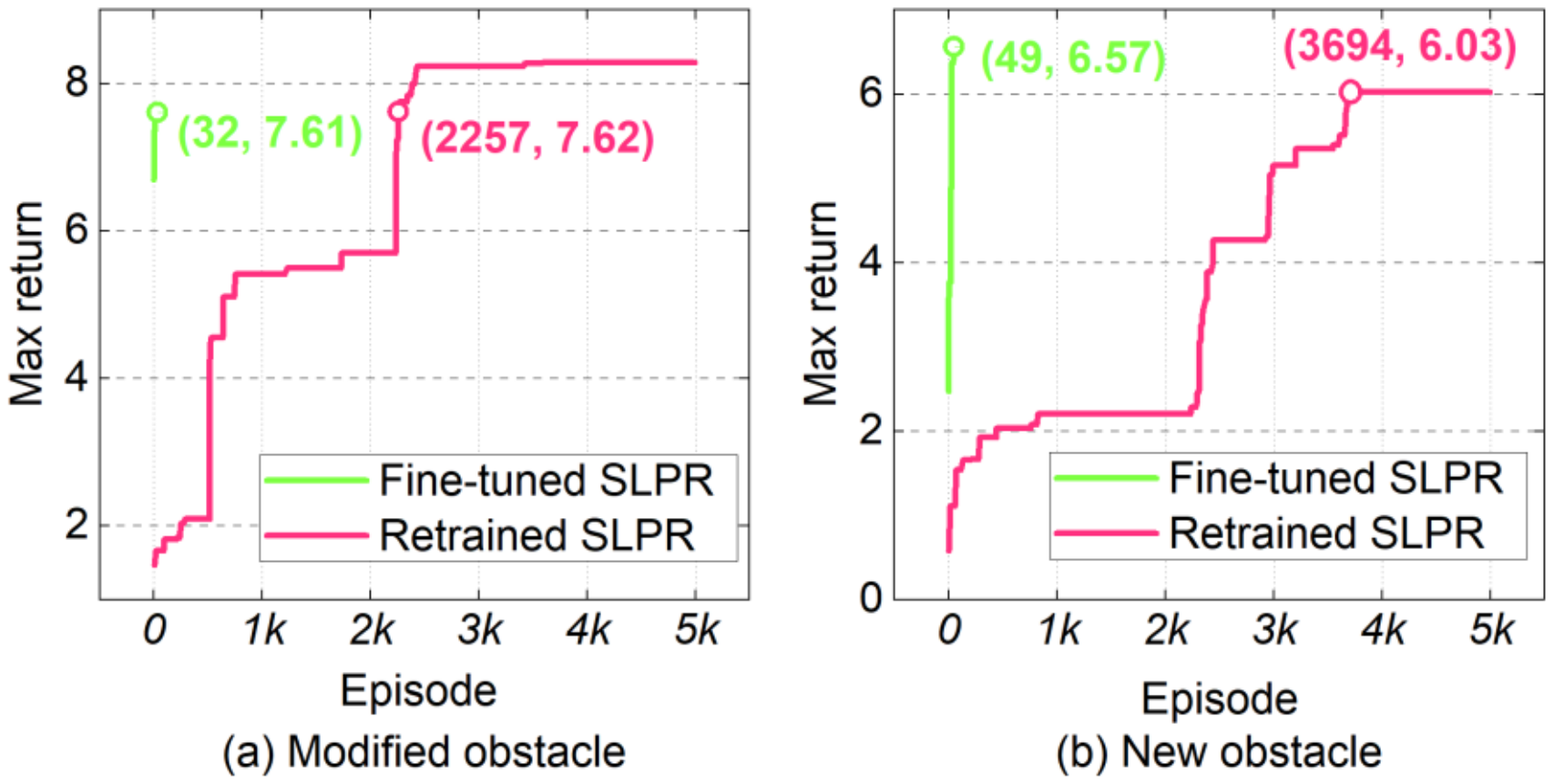

**Figure 13** Maximum return curves for fine-tuned and retrained SLPR. (a) In the “Modified obstacle” scenario; (b) In the “New obstacle” scenario.

The performance comparison of SLPR against the baseline methods in dynamic environments is presented in Table 4. Both BSRRT* and QPSO were retrained from scratch to adapt to the modified scenarios. The results highlight the significant computational advantage of the proposed approach. The adaptation latency of the SLPR agent was measured at only 24.18 seconds and 25.22 seconds for the two scenarios, respectively. Remarkably, this brief fine-tuning period yielded substantial performance gains. Within 30 seconds of adaptation, SLPR enabled successful pathfinding in the “New Obstacle” scenario and produced a significantly improved layout in the “Modified Obstacle” scenario. In contrast, the non-learning-based methods, BSRRT* and QPSO, required over 700 seconds to compute a new solution from scratch. Moreover, the pipe lengths generated by SLPR were considerably shorter than those obtained by BSRRT* and QPSO, and no violation points were observed in SLPR's results.

**Table 4** Performance of SLPR and several comparative methods in a dynamic environment

| Scenarios | Metrics | SLPR (old) | SLPR (fine-tune) | BSRRT* (retrained) | QPSO (retrained) |
|---|---|---|---|---|---|
| New obstacle | Length (mm) | 683.51 | 1003.47 | 2035.47 | 3069.76 |
| | Num of violation points | 0 | 0 | 0 | 0 |
| | Agent training time (s) | / | 24.18 | 1030.39 | 2222.09 |
| | Successful times | 0 | 1 | 1 | 1 |
| Modified obstacle | Length (mm) | 957.32 | 877.71 | 868.10 | 2409.96 |
| | Num of violation points | 0 | 0 | 18 | 0 |
| | Agent training time (s) | / | 25.22 | 706.69 | 1294.81 |
| | Successful times | 1 | 1 | 1 | 1 |

In summary, the proposed SLPR framework demonstrates a strong capability for rapid adaptation to design modifications via efficient fine-tuning. It eliminates the need for time-consuming re-optimization, substantially reducing the design person-hours required. SLPR not only performs effectively in static environments but also adapts swiftly to dynamic environmental changes, offering a robust and intelligent solution for aeroengine pipe routing.

## 6 Conclusion

This study presents a self-learning pipe routing (SLPR) framework for the automated design of free-form pipes in aeroengines. The pipe paths are represented using NURBS curves, which inherently ensure $C^2$ continuity, offering the necessary smoothness and geometric flexibility for navigating complex spatial environments. The SLPR framework constructs a potential energy lookup map that encodes various layout rules into a discretized spatial grid. By employing an efficient indexing function, it seamlessly integrates continuous-space path exploration with discrete-space rule evaluation, enabling near-instantaneous fitness assessment for any point along a candidate path. For path optimization, SLPR utilizes the learning-based PPO algorithm. Through continuous interaction with the design environment, the agent incrementally refines its routing policy and accumulates reusable knowledge of layouts. When design modifications occur, the agent can be rapidly fine-tuned to adapt to the new environment, thereby avoiding the computationally intensive retraining process required by conventional methods. The effectiveness of the proposed framework was validated through a comprehensive set of experiments. The primary findings are summarized as follows:

(1) Compared to traditional constant-curvature pipe routing, the SLPR-generated free-form pipe paths exhibit distinct advantages in the constrained annular layout spaces of aeroengines. They consistently reduce overall pipe length and geometric complexity, which is critical for achieving lightweight design and simplifying manufacturing.

(2) In comparative experiments against diverse algorithmic paradigms, including SAC, BSRRT*, and QPSO, SLPR achieved the shortest average path length of 670.48 mm, a 100% success rate, zero rule violations, and the highest computational efficiency.

(3) The SLPR framework demonstrated robust adaptability to design modifications through fine-tuning, achieving effective adaptation in approximately 25 seconds. This significantly reduces the design cycle time by eliminating the need for complete retraining. Moreover, the fine-tuned agent converged more quickly to superior solutions than those obtained via retraining from scratch.

In summary, the SLPR framework provides an efficient, intelligent, and adaptable solution for automated pipe routing. By addressing the requirements of lightweight, precision, and sustainability in modern aeroengine design, it holds substantial promise for real-world engineering applications.

Although this study successfully validates the applicability of the SLPR framework in complex single-path routing scenarios, it has not yet addressed the challenges associated with high-density, multi-pipe collaborative layouts. Furthermore, a gap remains between the current simulation-based framework and a fully validated, industrial-grade solution. Future work will focus on three key areas: (1) enhancing the core optimization process by incorporating manufacturability and physics-based constraints; (2) achieving collaborative optimization among multiple pipe layouts through multi-agent reinforcement learning and multi-objective optimization techniques; and (3) enabling seamless integration with industrial workflows and compliance with certification standards. Collectively, these efforts will advance the SLPR framework from a promising proof-of-concept to a robust, certifiable tool suitable for real-world aerospace engineering applications.

## Acknowledgements

This paper is funded by the National Natural Science Foundation of China (52275274), the Joint Funds of the National Natural Science Foundation of China (U20A20287), the Pioneer and Leading Goose R&D Program of Zhejiang, China (2024C01197), the Public Welfare Technology Application Projects of Zhejiang Province, China (LGG22E050008), the State Key Laboratory of Intelligent Manufacturing Equipment and Technology Open Project, China (IMETKF2024007). Zhejiang Province leading innovative entrepreneurial team project (2022R01012).

The authors thank Dave Rutland for sharing the Orenda 9/10 jet engine CAD model on GrabCAD.

## References

1 Yin Y H, Xu L D, Bi Z M, et al. A novel human-machine collaborative interface for aero-engine pipe routing. IEEE Trans Ind Inf, 2013, 9: 2187-2199

2 Chen Z S, Guan H, Yuan X L, et al. Rule-based generation of hvac duct routing. Autom Constr, 2022, 139: 14

3 Park J-H, Storch R L. Pipe-routing algorithm development: Case study of a ship engine room design. Expert Syst Appl, 2002, 23: 299-309

4 Fang F, Yu J, Xiong J, et al. Multi-solution pipe-routing method for the aeroengine with route constraints based on multi-objective optimization. Frontiers of Mechanical Engineering, 2024, 19: 37 5 Liu C, Wu L, Li G, et al. Ai-based 3d pipe automation layout with enhanced ant colony optimization algorithm. Autom Constr, 2024, 167: 105689

6 Zhang Z, Wu J, Liang B, et al. A new strategy for acquiring the forming parameters of a complex spatial tube product in free bending technology. J Mater Process Technol, 2020, 282: 116662

7 Wang Z, Lin Y, Qiu L, et al. Spatial variable curvature metallic tube bending springback numerical approximation prediction and compensation method considering cross-section distortion defect. Int J Adv Manuf Technol, 2022, 118: 1811-1827

8 Mazaheri H, Goli S, Nourollah A. A survey of 3d space path-planning methods and algorithms. ACM Comput Surv, 2025, 57: 32

9 Qu Y F, Jiang D, Gao G Y, et al. Pipe routing approach for aircraft engines based on ant colony optimization. J Aerosp Eng, 2016, 29: 10

10 Liu L X, Wang X, Yang X, et al. Path planning techniques for mobile robots: Review and prospect. Expert Syst Appl, 2023, 227: 30

11 Zhang Z, Zhang Y, Han R, et al. A generalized continuous collision detection framework of polynomial trajectory for mobile robots in cluttered environments. IEEE Rob Autom Lett, 2022, 7: 9810-9817

12 Yao Z, Zhang W, Shi Y, et al. Reinforcedrimjump: Tangent-based shortest-path planning for two-dimensional maps. IEEE Trans Ind Inf, 2020, 16: 949-958

13 Liu C, Wu L, Huang X, et al. Improved dynamic adaptive ant colony optimization algorithm to solve pipe routing design. Knowl-Based Syst, 2022, 237: 107846

14 Ren L, Fan X, Cui J, et al. A multi-agent reinforcement learning method with route recorders for vehicle routing in supply chain management. IEEE Trans Intell Transp Syst, 2022, 23: 16410-16420

15 Ma K, Liao S, Niu Y. Connected vehicles' dynamic route planning based on reinforcement learning. Future Gener Comput Syst, 2024, 153: 375-390

16 Elguea-Aguinaco Í, Serrano-Muñoz A, Chrysostomou D, et al. A review on reinforcement learning for contact-rich robotic manipulation tasks. Rob Comput Integr Manuf, 2023, 81: 102517

17 He Z, Dong L, Sun C, et al. Asynchronous multithreading reinforcement-learning-based path planning and tracking for unmanned underwater vehicle. IEEE Trans Syst Man Cybern: Syst, 2022, 52: 2757-2769

18 Chu Z, Wang Y, Zhu D. Local 2-d path planning of unmanned underwater vehicles in continuous action space based on the twin-delayed deep deterministic policy gradient. IEEE Trans Syst Man Cybern: Syst, 2024, 54: 2775-2785

19 Schulman J, Wolski F, Dhariwal P, et al. Proximal policy optimization algorithms. 2017.

20 Liu C, Wu L, Li G X, et al. Improved multi-search strategy a* algorithm to solve three-dimensional pipe routing design. Expert Syst Appl, 2024, 240: 26

21 Lin Y, Zhang Q. A multi-objective cooperative particle swarm optimization based on hybrid dimensions for ship pipe route design. Ocean Eng, 2023, 280: 114772

22 Zhang Q, Lin Y. Integrating multi-agent reinforcement learning and 3d a* search for facility layout problem considering connector-assembly. J Intell Manuf, 2024, 35: 3393-3418

23 Qu Y F, Jiang D, Zhang X L. A new pipe routing approach for aero-engines by octree modeling and modified max-min ant system optimization algorithm. J Mech, 2018, 34: 11-19

24 Elbanhawi M, Simic M, Jazar R. Randomized bidirectional b-spline parameterization motion planning. IEEE Trans Intell Transp Syst, 2016, 17: 406-419

25 Kawashima S, Itoh H, Kawamura Y. Calculation of the number of ship collision candidates using mesh-based estimation method for ship traffic data. J Mar Sci Technol, 2022, 27: 1233-1251

26 Liu Y, Zhao W, Sun R, et al. Optimal path planning for automated dimensional inspection of free-form surfaces. J Manuf Syst, 2020, 56: 84-92

27 Yan C, Shi Y, Li Z, et al. Chapter 2 - software algorithm and route planning. In: C. Yan, Y. Shi, Z. Li, S. Wen, Q. Wei (Eds.) Selective laser sintering additive manufacturing technology, Academic Press; 2021, p. 123-249.

28 Xu Y, Tong X, Stilla U. Voxel-based representation of 3d point clouds: Methods, applications, and its potential use in the construction industry. Autom Constr, 2021, 126: 103675

29 Ito T. A genetic algorithm approach to piping route path planning. J Intell Manuf, 1999, 10: 103-114

30 Dong Z, Bian X. Ship pipe route design using improved a* algorithm and genetic algorithm. IEEE Access, 2020, 8: 153273-153296

31 Cetinsaya B, Reiners D, Cruz-Neira C. From pid to swarms: A decade of advancements in drone control and path planning - a systematic review (2013–2023). Swarm Evol Comput, 2024, 89: 101626

32 Liu S, Jiang H, Chen S, et al. Integrating dijkstra's algorithm into deep inverse reinforcement learning for food delivery route planning. Transp Res E: Logist Transp Rev, 2020, 142: 102070

33 Lin Z, Wu K, Shen R, et al. An efficient and accurate a-star algorithm for autonomous vehicle path planning. IEEE Trans Veh Technol, 2024, 73: 9003-9008

34 Qi J, Yang H, Sun H. Mod-rrt*: A sampling-based algorithm for robot path planning in dynamic environment. IEEE Trans Ind Electron, 2021, 68: 7244-7251

35 Sandurkar S, Chen W. Gaprus—genetic algorithms based pipe routing using tessellated objects. Comput Ind, 1999, 38: 209-223

36 Wang C, Wang Z, Zhang S, et al. Adam-assisted quantum particle swarm optimization guided by length of potential well for numerical function optimization. Swarm Evol Comput, 2023, 79: 101309

37 Qu Y, Jiang D, Yang Q. Branch pipe routing based on 3d connection graph and concurrent ant colony optimization algorithm. J Intell Manuf, 2018, 29: 1647-1657

38 Haarnoja T, Zhou A, Abbeel P, et al. Soft actor-critic: Off-policy maximum entropy deep reinforcement learning with a stochastic actor. In: Proceedings of the 35th International Conference on Machine Learning, 2018. 1861-1870.

39 Al-Hilo A, Samir M, Assi C, et al. Uav-assisted content delivery in intelligent transportation systems-joint trajectory planning and cache management. IEEE Trans Intell Transp Syst, 2021, 22: 5155-5167

40 Xi M, Yang J, Wen J, et al. An information-assisted deep reinforcement learning path planning scheme for dynamic and unknown underwater environment. IEEE Trans Neural Networks Learn Syst, 2023, 1-12

41 Yang J, Huo J, Xi M, et al. A time-saving path planning scheme for autonomous underwater vehicles with complex underwater conditions. IEEE Internet Things J, 2023, 10: 1001-1013

42 Kim Y, Lee K, Nam B, et al. Application of reinforcement learning based on curriculum learning for the pipe auto-routing of ships. J Comput Des Eng, 2023, 10: 318-328

43 Li X, Zhao H, He X, et al. A novel cartesian trajectory planning method by using triple nurbs curves for industrial robots. Rob Comput Integr Manuf, 2023, 83: 102576

44 Liu Y, Chen C, Wang Y, et al. A fast formation obstacle avoidance algorithm for clustered uavs based on artificial potential field. Aerosp Sci Technol, 2024, 147: 108974

45 Rutland D. Orenda 9/10 jet engine almost complete assembly [CAD model]. GrabCAD, n.d. https://grabcad.com/library/orenda-9-10-jet-engine-almost-complete-assembly-1

46 Akiba T, Sano S, Yanase T, et al. Optuna: A next-generation hyperparameter optimization framework. In: Proceedings of the 25th ACM SIGKDD International Conference on Knowledge Discovery & Data Mining, Anchorage, AK, USA, 2019. 2623–2631.

47 Feng H, Hu Q, Zhao Z, et al. Smooth path planning under maximum curvature constraints for autonomous underwater vehicles based on rapidly-exploring random tree star with b-spline curves. Eng Appl Artif Intell, 2024, 133: 108583